\documentclass{article}

\usepackage{times}
\usepackage{latexsym}
\usepackage{graphicx}
\usepackage{multirow}
\usepackage{amsmath}
\usepackage{lipsum}
\usepackage[]{mdframed}

\usepackage{microtype}
\usepackage{subfigure}
\usepackage{booktabs} %
\usepackage{array}    
\usepackage{longtable}
\usepackage{float}
\usepackage{adjustbox} 
\usepackage{amssymb} 
\usepackage{pifont} 
\newcommand{\cmark}{\ding{51}}%
\newcommand{\xmark}{\ding{55}}%

\usepackage{hyperref}

\usepackage[accepted]{icml2024}

\usepackage{mathtools}
\usepackage{amsthm}

\usepackage[capitalize,noabbrev]{cleveref}

\theoremstyle{plain}

\theoremstyle{definition}

\theoremstyle{remark}

\usepackage[textsize=tiny]{todonotes}

\icmltitlerunning{Text Serialization and Their Relationship with the Conventional Paradigms of Tabular Machine Learning}

\begin{document}

\twocolumn[
\icmltitle{Text Serialization and Their Relationship with the Conventional Paradigms of Tabular Machine Learning}



\icmlsetsymbol{equal}{*}

\begin{icmlauthorlist}
\icmlauthor{Kyoka Ono}{comp,zzz,equal}
\icmlauthor{Simon A. Lee}{yyy,equal}
\end{icmlauthorlist}

\icmlaffiliation{yyy}{Department of Computational Medicine
University of California, Los Angeles
Los Angeles, California, USA 90095 }
\icmlaffiliation{comp}{Department of Statistics and Data Science, University of California, Los Angeles}
\icmlaffiliation{zzz}{Department of Natural Sciences, International Christian University, Mitaka, Tokyo, Japan}

\icmlcorrespondingauthor{Simon Lee}{simonlee711@g.ucla.edu}

\icmlkeywords{Machine Learning, ICML}

\vskip 0.3in
]



\printAffiliationsAndNotice{\icmlEqualContribution} 

\begin{abstract}
Recent research has explored how Language Models (LMs) can be used for feature representation and prediction in tabular machine learning tasks. This involves employing text serialization and supervised fine-tuning (SFT) techniques. Despite the  simplicity of these techniques, significant gaps remain in our understanding of the applicability and reliability of LMs in this context. Our study assesses how emerging LM technologies compare with traditional paradigms in tabular machine learning and evaluates the feasibility of adopting similar approaches with these advanced technologies. At the data level, we investigate various methods of data representation and curation of serialized tabular data, exploring their impact on prediction performance. At the classification level, we examine whether text serialization combined with LMs enhances performance on tabular datasets (e.g. class imbalance, distribution shift, biases, and high dimensionality), and assess whether this method represents a state-of-the-art (SOTA) approach for addressing tabular machine learning challenges. Our findings reveal current pre-trained models should not replace conventional approaches.
\end{abstract}

\section{Introduction}

In the field of natural language processing (NLP), a paradigm shift has occurred, driven by the emergence of Language Models (LM) technologies rooted in the transformer architecture \citep{vaswani2017attention}. These advancements have led to immense progress across various domains of machine learning (ML) and artificial intelligence (AI). Leveraging sophisticated techniques such as transfer learning \citep{weiss2016survey} and attention mechanisms \citep{bahdanau2014neural}, LMs have demonstrated exceptional capabilities in tasks encompassing language understanding \citep{devlin2018bert}, translation \citep{lewis2019bart}, and text generation \citep{radford2018improving}, thereby significantly influencing applications within the field of NLP. However, researchers from various fields have discovered that these LMs are not limited to conventional tasks. Consequently, there has been a surge of research into other areas and domains, such as question-answering \citep{radford2019language,su2019generalizing} and mathematical reasoning \citep{trinh2024solving,wang2023mathcoder,imani2023mathprompter}, among others.

Therefore, in this paper, we focus on the ability of LMs to solve tabular machine learning tasks as introduced by \citep{hegselmann2023tabllm, sahakyan2021explainable, dinh2022lift, fang2024large}. These studies utilize text serialization—converting tabular data into natural language representations—combined with supervised fine-tuning (SFT) to evaluate LMs' capability on supervised machine learning tasks. Yet, current papers do not explore whether this process or these LMs could represent a state-of-the-art (SOTA) approach in machine learning. This oversight is especially significant in light of previous assertions that gradient boosting methods outperform deep learning strategies \citep{grinsztajn2022tree}.

These previous works also did not determine whether various data curation measures are required for obtaining accurate results and how to adequately handle the common data preparation practices commonly used in tabular machine learning (e.g. missing data, feature scaling, etc.). As a result, there are open questions in the current literature about text serialization and whether they align with conventional machine learning paradigms.

In this work, we explore the unresolved questions related to text serialization. We believe this research is crucial for contrasting the differences between traditional ML methods and emerging methodologies like ``text serialization'' developed for LM technologies. Thus, we rigorously analyze numerous publicly available tabular datasets and detail the various experiments conducted to gain insights into the current questions in this area of research. We aim to determine whether data curation is necessary and assess whether these pre-trained LMs should be used over traditional tabular solvers like gradient boosting under varying dataset characteristics. The contributions of this paper are as follows:

\begin{itemize}
    \vspace*{-0.3cm}
    \item We investigate whether open-source LMs, in conjunction with text serialization, can achieve state-of-the-art (SOTA) performance compared to current ML methods in supervised learning tasks. We aim to determine whether pre-trained models should be preferred over previously established gradient-boosted methods.
    \vspace*{-0.3cm}
    \item We investigate how various data curation strategies for text serialization, such as addressing missing values, feature importance, and feature scaling, affect prediction performance. We also consider whether these common protocols should be followed for language modeling. 
    \vspace*{-0.3cm}
    \item We investigate the adaptability and generalization capabilities of LMs across different characteristics of tabular datasets that are commonly encountered in real world datasets (e.g. high dimensionality, imbalance).
    \vspace*{-0.2cm}
    \item We evaluate the robustness of LM-based models against common distribution shifts and dataset biases, examining how their pretrained parameters respond to these characteristics.
    \vspace*{-0.3cm}
\end{itemize}



\section{Related Works}

\subsection{Text Serialization}

Text Serialization introduced by \cite{hegselmann2023tabllm, dinh2022lift,gidroltext, jaitly2023better, lee2024enhancing} created an interface to allow an easy integration with tabular data to LMs by converting tabular data fields into a natural language representations. Since its emergence there have been numerous papers in various applications including healthcare that have adopted a similar approach \cite{chen2024multimodal,kim2024health,hegselmann2024data,belyaeva2023multimodal}. Lee et al. found that text serialization, in particular, proved effective for handling categorical tabular data with a large number of classes. They observed that a natural language representation outperformed engineered features like one-hot encoding \citep{lee2024multimodal}. Text serialization has found application in various reasoning tasks, such as feature extraction, enabling systems to extract information from tables or databases to answer queries, as seen in Question and Answer (Q\&A) scenarios \cite{min2024exploring, sui2024table, li2024can}.

Following this conversion from tabular to text, the resulting data can be directly input into foundation models (e.g., BERT \cite{devlin2018bert}, GPT \cite{brown2020language}, etc.) to obtain rich feature representations in the form of high-fidelity vectors. Recent research has focused extensively on representing numerical data \cite{gorishniy2022embeddings, golkar2023xval}, where these foundation models have demonstrated competitive and often superior performance compared to current models like XGBoost \cite{chen2016xgboost} and LGBM \cite{ke2017lightgbm}, showing recent evidence against previous claims of boosted methods being the SOTA \cite{grinsztajn2022tree}.

\subsection{Tabular Deep Leaning}

Deep learning has emerged as an exceptional computational framework across numerous disciplines due to its ability to learn complex patterns in large datasets \cite{zhang2018analyzing, feng2019fringe}, generalize effectively \cite{sanh2021multitask}, apply transfer learning techniques \cite{torrey2010transfer, zhuang2020comprehensive, pan2009survey, niu2020decade, levin2022transfer}, and scale with powerful hardware \cite{mayer2020scalable, chilimbi2014project, rouhani2018deepsecure}. Tabular deep learning has been investigated over the years, yet there remains no consensus on whether it represents the optimal modeling approach for this type of data \cite{shwartz2022tabular, borisov2022deep, gorishniy2021revisiting}. Despite this lack of consensus, many groups continue to explore this field extensively. Examples include TabNet \cite{arik2021tabnet}, TabPFN \cite{hollmann2022tabpfn}, SAINT \cite{somepalli2021saint}, TabTransformer \cite{huang2020tabtransformer}, NODE \cite{popov2019neural}, and TaBERT \cite{yin2020tabert}. Kadra et al. demonstrated that even simple neural nets can produce high-performing models compared to baselines \cite{kadra2021well}.

\begin{figure*}[h!]
    \centering
    \includegraphics[width=6.5in]{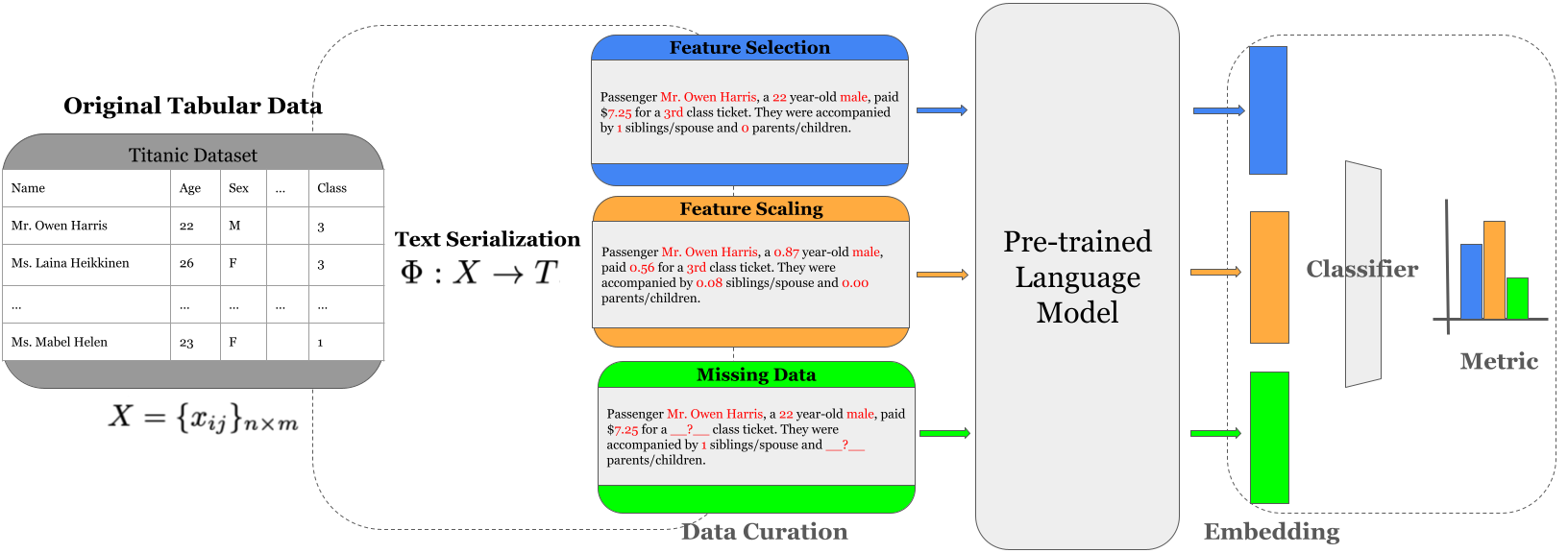}
    \caption{\textbf{Data Curation In Text Serialization:} In this work we explore several common data curation strategies used in tabular machine learning and determine whether these practices should be adopted in Language Model technologies.}
    \label{arc}
\end{figure*}

More recently, there has been a resurgence of interest in tabular deep learning, driven by advancements in Language Model (LM) technology. Notably, models like TabLLM \cite{hegselmann2023tabllm}, LIFT \cite{dinh2022lift}, MEME \cite{lee2024multimodal, lee2024emergency}, and others \cite{zhang2023towards} have showcased robust performance in both few-shot and fully trained scenarios. However, when evaluating these LMs with zero or few shots, it's challenging to determine whether they are learning the task \cite{webson2021prompt} or merely hallucinating based on simpler classification tasks, which complicates model evaluation \cite{ji2023towards, lee2024large}. Nevertheless, fine-tuning these language models enables them to be adapted to perform specific tasks using a few-shot (minimal data) approach \cite{harari2022few, liu2022few, perez2021true, zhao2021calibrate}.

Despite recent advances in language models and tabular machine learning, numerous unanswered questions remain regarding the use of language models in this field. Therefore, this study aims to comprehensively address some knowledge gaps concerning the systematic approach to the machine learning pipeline and how these new approaches align with conventional paradigms. Additionally, we would like to highlight and address other common scenarios where pre-trained language models can be beneficial and whether these general models should be adopted and surpass previously state-of-the-art models that are primarily based on gradient boosting. We hypothesize that language models do not adhere to conventional paradigms and do not require data curation techniques, but we believe that these pre-trained models can be effective tabular solvers.


\section{Methodology}

\subsection{Text Serialization}
\textbf{Problem Formulation:} Text serialization is the process of transforming structured tabular data \( X \) with dimensions \( n \times m \) into textual representations. Here, \( n \) is the number of samples, and \( m \) represents the number of features. In the study by Hegselmann et al. on TabLLM, they identified that using text templates and list readouts provided the best results among various serialization strategies \cite{hegselmann2023tabllm}. Therefore we will adopt a text template approach within our analysis. Mathematically, this transformation can be represented as follows: Let \( X = \{x_{ij}\}_{n \times m} \) be the input dataset, where \( x_{ij} \) is the value of the \( i \)-th sample in the \( j \)-th feature. Let \( Y = \{y_i\}_n \) be the corresponding set of labels for each sample in \( X \). The goal of text serialization is to define a mapping \( \Phi: X \rightarrow T \), where \( T = \{t_i\}_n \) represents the serialized text derived from the data in \( X \). This mapping function \( \Phi \) uses  template filling to convert \( x_i \) into the corresponding serialized text \( t_i \). From this textual representation, we will utilize this data alongside the labels for our supervised fine-tuning in classification tasks.
\subsection{Language Model Selection}

In our study, we need to select a language model backbone that we will use in the study. There are numerous backbones to choose from, but we filter these by selecting a language model that was pretrained on text classification objectives. Therefore, in order to select the best Language Model (LM) for our benchmark, we will conduct an evaluation on multiple open source LMs sourced from the huggingface sequence classification library \citep{wolf2019huggingface}. We additionally benchmark several models from the Massive Text Embedding Benchmark (MTEB) \cite{muennighoff2022mteb}—a comprehensive framework designed to evaluate the performance of text embedding models across a wide range of tasks—and select models based on their rank in text classification. This is in an effort to find the LM that provides the best representation for our serialized textual data. In Table~\ref{TableModel}, we highlight which LMs we evaluate with a short description describing each of them. 

\subsection{Current Understanding and Limitations}
\subsection*{What do we know about Text Serialization}

From the literature, there are concrete findings that have been made in text serialization. Text serialization has enabled the integration of tabular data with language models (LMs), leading to competitive performance in datasets with minimal samples (few-shot) \cite{hegselmann2023tabllm, yang2024smalltolarge} or no samples at all (zero-shot) \cite{wei2021finetuned, kojima2022large, zhong2021adapting}. This success is due to converting data into a natural language format, which allows for the effective application of transfer learning using hundreds of millions of pre-trained parameters within a LM to carry out inference. While recent works have progressed towards the ability to read structured data \cite{song2023restgpt, chen2024beyond, yao2023sai}, text serialization appears to remain the best method for integrating tabular data with LMs. Another use case of text serialization was identified when tabular data has categorical data with a high number of classes or heterogenous data (numerical, categorical, free text) within the tabular fields \cite{lee2024multimodal}. This methodology allows us to seamlessly preserve all the data in its natural form (no feature engineering necessary), represented all as text. Groups including \cite{belyaeva2023multimodal, chen2024multimodal} also demonstrated that text serialization was particularly effective when integrated with paired multimodal datasets, enabling contrastive methods to shared latent representations \cite{radford2021learning}.

\subsection*{What needs to be addressed}
\vspace{-0.1cm}
While considerable progress has been made in advancing tabular data with LM  technologies, many intermediate steps at both the data and classification levels remain undisclosed. This paper aims to address some of the key gaps in the current literature, providing a more comprehensive understanding of the existing challenges and solutions.

\paragraph{Data Questions:} Many questions remain regarding whether text serialization or LMs adhere to similar approaches as those of traditional machine learning paradigms. This is particularly relevant in the data curation process when handling raw data that contains missing values, the need to identify important and unimportant features, and dealing with differently distributed numerical data. Applying data curation is often a crucial component in traditional machine learning pipelines, but no study has yet examined whether similar approaches are required in LM technologies for supervised tasks. A visualization of this exploration can be seen in Figure \ref{arc}.

\paragraph{Classification Questions:} In addition, there have been no studies regarding whether pre-trained LMs should be used for all tabular supervised classification tasks. Therefore, we explore several datasets with commonly encountered characteristics and benchmark them against various tabular SOTA models and traditional machine learning methods. We aim to determine whether LMs support or contradict previous claims that gradient boosting performs better than deep learning-based models in tabular tasks \citep{grinsztajn2022tree}. 

\section{Experimental Setup}

\subsection{Data}
\label{data}

In our study, we utilize eight datasets, which we divide into two groups. 

\begin{table}[h!]
  \small 
  \centering
  \caption{The Dataset and its Characteristics used in the analysis. The $\heartsuit$ denotes that there is missing data. The $\dagger$ denotes a distribution shift dataset. The $\diamondsuit$ denotes an imbalanced dataset. \cite{gardner2023tableshift}. The $\clubsuit$ denotes that these are well-documented bias datasets with high number of classes.  The $\spadesuit$ denotes a dataset with a large number of features. We also indicate whether the datasets are binary or not. If not binary they are considered multi-class.}
  
  \begin{adjustbox}{width=\columnwidth}
  \begin{tabular}{lcccc}
    \hline
    \textbf{Dataset} & \textbf{Sample Size ($n$)} & \textbf{\# of Features ($m$)} & \textbf{Binary} \\
    \hline
    Iris   & $150$ & $4 $ & \xmark       \\
    Diabetes & 784 & 8 & \cmark\\
    Titanic $\heartsuit$ & $891$ & $11$&  \cmark          \\
    Wine    & $178$ & $13$  &  \xmark      \\
    HELOC$^\dagger$ & $10,459$  & $23$ & \cmark   
    \\
        Fraud$^{\diamondsuit}$    & $284,807$ & $30$  &\cmark       \\
    Crime$^\clubsuit$    & $878,049$ & $8$ &   \xmark      \\
    Cancer$^\spadesuit$ & $801$ & $20,533$ & \xmark\\
    \hline
  \end{tabular}
  \end{adjustbox}
  \label{charact}
\end{table}

\paragraph{Baseline Datasets:} The first four datasets are commonly used baselines in tabular machine learning. These datasets include the IRIS, Wine, Diabetes, and Titanic Dataset, which are either binary or multiclass (3 class) classification problems sourced from the UCI data repository or previous literature \citep{asuncion2007uci, smith1988using}. We utilize these baseline datasets in our data-level experiments to identify which preprocessing steps affect relative performance and should be adapted for our SOTA experiments.

\paragraph{Experimental Datasets:} The second group of datasets can be labeled as a set of datasets with interesting and common machine learning characteristics. We utilize these datasets only in our SOTA evaluation by using the identified preprocessing steps from our previous experiments. These datasets include an Identifying Targets for Cancer Using Gene Expression Profiles dataset, which includes high dimensionality \citep{misc_gene_expression_cancer_rna-seq_401}; the HELOC Dataset \cite{brown2018heloc}, which contains well documented distribution shift identified by \citep{gardner2023tableshift}; the San Francisco Crime dataset, which contains inherent biases towards certain neighborhoods \citep{asuncion2007uci}; and the Credit Card Fraud dataset, which contains class imbalance \citep{dal2015calibrating} (0.172\% of data is fraud). These datasets contain a mixture of binary and multi-class classification tasks. All characteristics to the datasets including sample size and feature size can be found in Table \ref{charact}. Additional details about what the raw data looks like and how we serialized it in different ways can be found in Appendix Section \ref{dataset}.

\subsection{Experiments}

At the data level, we've identified gaps in the literature related to data curation for text serialization and whether they follow approaches similar to traditional machine learning paradigms. To explore the effects of various preprocessing measures on serialized tabular data, we utilize a baseline model, where no data curation is performed. We then explore how applying different preprocessing techniques, affect performance relative to the baseline.

Additionally, at the classification level, we are interested in testing the robustness of LMs when faced with commonly encountered real-life dataset characteristics. With the identified requisite from our data curation experiments, we evaluate the LMs against existing methods and commonly used ML methods on datasets that exhibit class imbalance, distribution shift among other. By introducing these challenges into our benchmark datasets, we aim to evaluate the relative performance of LMs in tackling fundamentally difficult challenges in tabular machine learning. We detail our experiments in greater detail in the proceeding subsections.

\subsection*{Data Experiments}
\paragraph{Feature Selection} Feature selection is the process of identifying and selecting a subset of relevant features from the original set of features to improve model performance and efficiency \cite{guyon2003introduction}. In our first experiment, we compare a baseline model, where no feature selection is applied, to a model where feature selection is utilized. We employ two feature selection methods: one using SHapley Additive exPlanations (SHAP) values extracted from an XGBoost model and another using the ANOVA F-test \cite{st1989analysis}. Further details on how these features are derived from these methods can be found in Appendix Section \ref{FS}. We then assess whether feature selection yields better, worse, or nuanced results. We further include serialized text in the appendix to give readers a view what these sentences look like with and without feature selection attached in Appendix Section \ref{dataset}.
\vspace{-0.35cm}

\paragraph{Feature Scaling \& Outlier Handling} \label{fss} Feature scaling involves converting features within a dataset to ensure they are on a similar scale, thus preventing certain features from dominating others in the analysis. We explore standardizing features (subtract the mean ($\mu$) and divide by the standard deviation ($\sigma$)) when they are on different scales and the machine learning algorithm is scale-sensitive. We normalize features (rescale to the range [0, 1]) to bring all features to a common range, particularly in the presence of outliers. Additionally, we apply log transformation when the data is skewed or contains outliers, as it can help mitigate the impact of extreme values and make the distribution more normal. These measures are applied to the Titanic \cite{eaton1995titanic} datasets based on the characteristics of its dataset, and we report whether such steps are necessary.
\vspace{-0.35cm}
\paragraph{Missing Data Handling \& Imputation} Missing data handling and imputation involve techniques for addressing and filling in missing values within a dataset to ensure completeness and maintain the integrity of the analysis. Unlike in traditional tabular machine learning, a clear method for handling missing values remains unclear. Therefore, we explore the effects of ignoring missing values (equivalent to dropping that single cell) and adding filler sentence techniques as a form of imputation, similar to those described in \cite{lee2024multimodal}. We will then perform a sensitivity analysis observing how much the logarithm of odds (logits) for each class change based on these imputation strategies.

\subsection*{Classification Experiments}

\paragraph{SOTA Benchmarks on Various Tabular Datasets} In our classification experiments we are particularly interested in seeing how LM perform compared to traditional Machine learning models, and several models from the literature. We test for SOTA in all the baseline datasets as well as our experimental datasets referenced in Section \ref{data}. These include datasets with high dimensionality, distribution shift, bias, and class imbalance. We don't perform data corrections (e.g. SMOTE, etc.), and instead want to assess the performance with these included characteristics.

\subsection{Benchmarking Baseline Models}

To evaluate the relative performance of text serialization and SFT we identify models commonly used for tabular machine learning to include in the benchmark that have excelled at tabular tasks. The models included were sourced from \cite{dinh2022lift, hegselmann2023tabllm} in the evaluation include Support Vector Machines (SVM) with the Radial basis function (RBF) kernel \cite{cortes1995support}, Light Gradient boosted machines \cite{ke2017lightgbm}, and XGBoost \cite{chen2016xgboost}. From the literature we also use Tabnet \cite{arik2021tabnet} and TabPFN \cite{hollmann2022tabpfn} which were optimized on tabular tasks. The metrics we will use to evaluate models include: f1, accuracy, Area Under the Receiver Operating Characteristic (AUROC), and mathew's correlation coefficient (MCC) \cite{chicco2020advantages}. When classification objectives are not binary we include the macro averaging strategy to create a uniform view of performance metrics across all methods. 

\subsection{Training and Model Optimization}

In terms of optimizing the language model performance, we elect for a standard learning rate of 2e-4 with a learning rate scheduler to tune this parameter dynamically. We also include a dropout of 0.3 to ensure that these models are not overfitting during fine tuning. We elect for a batch size of 64 on each dataset. We minimize on the Binary Cross Entropy loss for binary classification, and Cross-entropy for multi-class classification. We evaluate our models using Pytorch and use LMs sourced from huggingface. We do all evaluations on a single Tesla V100 GPU with 16GB of VRAM.

In the standard machine learning models, we elect to conduct a five-fold cross-validation grid search to find optimal hyperparameters for the benchmark. We showcase these hyperparameters we searched for in the appendix for reproducibility purposes (Appendix Section \ref{hyperparemters}).

\section{Results}

\subsection{Language Model Evaulation}

\begin{figure}[h!]
    \centering
    \includegraphics[width=\columnwidth]{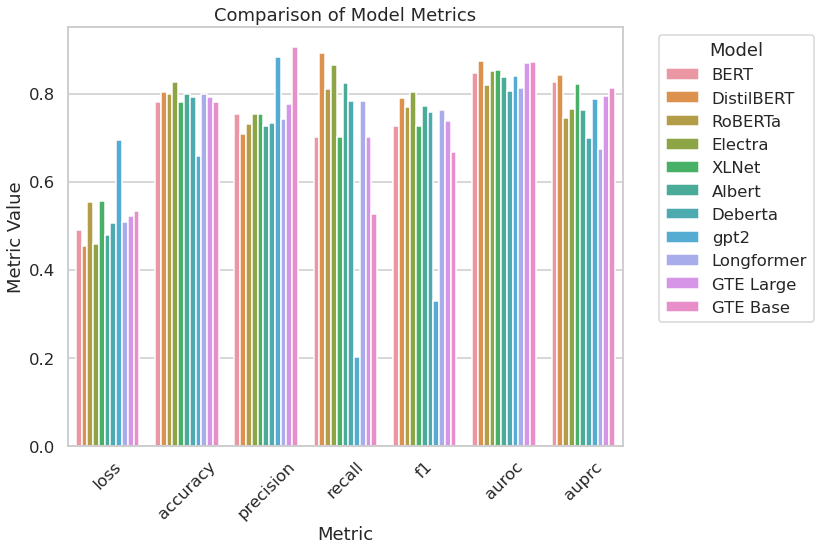}
    \caption{\textbf{Language Model Benchmark:} The evaluation of several LM backbones that are benchmarked to be our TabLM.}
    \label{MS}
\end{figure}

We begin our analysis by identifying our ``TabLM" through a benchmarking study on a set of language models sourced from the Huggingface sequence classification and Massive Text Embedding Benchmark (MTEB) \cite{muennighoff2022mteb}. We conducted this analysis using the Titanic baseline dataset and used  serialized text template inputs to conduct a proper evaluation. From Figure \ref{MS}, we find that DistilBERT is the best performing model, and we elect this to be our Tabular Language Model, which we will refer to as TabLM. One notable finding from this evaluation is that the MTEB's ranking of text classification is not compatible with tabular machine learning tasks, as evidenced by standard models outperforming the General Text Embedding (GTE) model \cite{li2023towards}. Additionally, the varying fluctuations across performance metrics illustrate how different pre-training objectives incorporated within these foundational models may optimize different performance metrics. Further details are located in (Section \ref{resultsss}).

\subsection{Data Curation Results}

\paragraph{Feature Selection}

In our feature selection experiment, we compare the performance between a baseline language model (LM) without feature selection and an LM that uses shorter serialized sentences containing only important features. These features are identified through XGBoost feature importance and visualized using SHapley Additive exPlanations (SHAP) values and ANOVA F-tests. 
\vspace{-0.1cm}
\begin{table}[h!]
\caption{Benchmark study with and without feature selection}
\label{r1}
\begin{adjustbox}{width=\columnwidth}
\begin{tabular}{l|cc|cc| c}
\toprule
Dataset & \multicolumn{2}{c|}{Without Feature Selection} & \multicolumn{2}{c}{With Feature Selection} & Improved? \\
\midrule
Metrics & AUROC & F1 & AUROC & F1 & \\
\midrule
Iris &  \textbf{1.000} & \textbf{1.000} & \textbf{1.000} & \textbf{1.000} & ---  \\
Wine & 0.952 & 0.944 & \textbf{0.976} & \textbf{0.972} & \cmark  \\
Diabetes & 0.654 & 0.621 & \textbf{0.659} & \textbf{0.659} & \cmark \\
Titanic $\heartsuit$ & \textbf{0.786} & \textbf{0.871} & 0.777 & 0.852 & \xmark  \\
\bottomrule
\end{tabular}
\end{adjustbox}
\end{table}

This study reveals that feature selection appears to have a positive effect on both F1 score and AUROC in most evaluation datasets, as seen in Table \ref{r1}. While the results are somewhat nuanced, we observe that selecting appropriate features for serialization tends to enhance performance in classification tasks and will likely be true in datasets with higher dimensionality.

\paragraph{Feature Scaling \& Outlier Handling} In our experiment on feature scaling and outlier handling, we benchmark models that serialize their numerical data using various feature scaling methods to compare their performance across multiple metrics. This evaluation specifically focuses on the Titanic dataset, which exhibits right-skewed distributions in both the \texttt{fare} and \texttt{age} features. To address these issues, we employ standardization, normalization, and logarithmic transformations on these features, applying corrections that offer different benefits as detailed in Section \ref{fss}. Each method is analyzed for its effectiveness in mitigating the impact of skewness and improving model performance, providing a comprehensive understanding of how feature scaling can influence key performance indicators.

\begin{figure}[h!]
    \centering
    \includegraphics[width=2.5in]{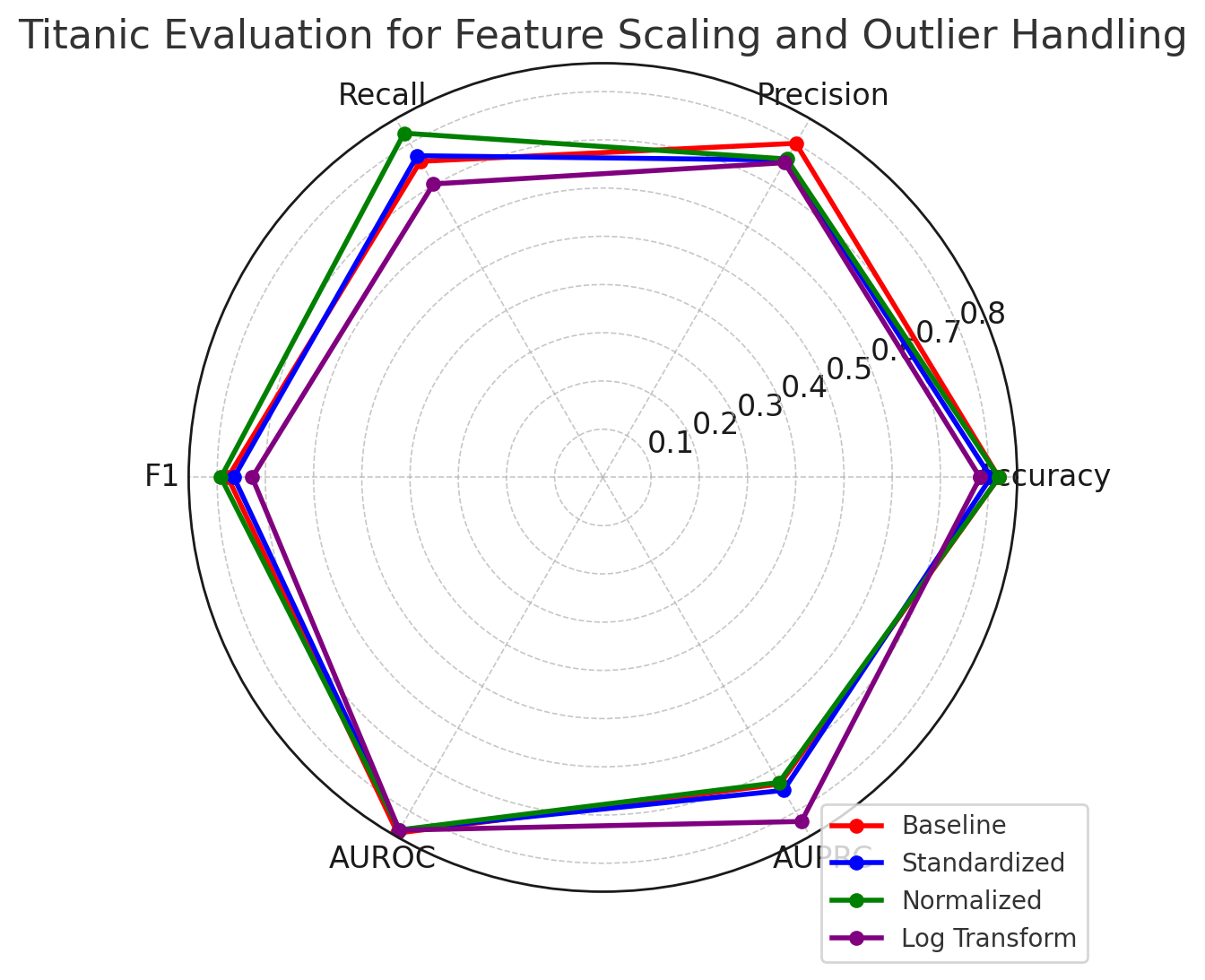}
    \caption{An evaluation using various feature scaling methods that include outlier handling. We see very nuanced results and these methods should be applied with some background knowledge of the dataset.}
    \label{shap}
\end{figure}
\vspace{-0.1cm}
Our analysis reveals nuanced results, where various feature scaling methods yield marginal gains and deficits. Based on these results, we advise that scaling methods should be applied in accordance with the classification objectives. However, it is also likely possible to achieve acceptable results without performing feature scaling.
\vspace{-0.2cm}

\begin{table*}[h!]
\caption{The benchmarking metrics, which include Accuracy, F1 Score, Area under the Receiver Operating Curve (AUROC), and Matthews Correlation Coefficient (MCC), were displayed. The results suggest that TabLM does not consistently outperform other methods on baseline datasets. Additionally, the findings indicate that there is no definitive method for handling tabular tasks.}
\vskip 0.15in
\begin{center}
\label{evaluation}
\begin{small}
\begin{sc}
\begin{adjustbox}{width=\textwidth}
\begin{tabular}{p{2cm}p{2.8cm}|cccc|c|c}
\toprule
\multicolumn{7}{c}{\textbf{State of the Art Evaluation - Baseline Datasets}} \\  
Dataset & Method & Accuracy & F1 & AUROC & MCC & Current State of the Art & TabLM SOTA? \\
\hline
\multirow{6}{*}{Iris} & SVM (RBF)&1.0000 &1.0000 &1.0000& 1.1870&\multirow{6}{*}{1.0000 (Acc)\cite{ojha2020multi}}&\multirow{6}{*}{\xmark}\\ 
& LGBM&1.0000 &1.0000 &1.0000& 1.1870&\\ 
& XGBoost&1.0000 &1.0000 &1.0000& 1.1870&\\ 
& TabNet&1.0000 &1.0000 &1.0000& 1.1870&\\
& TabPFN&1.0000 &1.0000 &---& 1.1870&\\ 
& TabLM &1.0000 &1.0000 &1.0000& 1.1870&\\ 
\hline
\multirow{6}{*}{Wine} & SVM (RBF)&0.8333&0.8107&0.9414&1.2004&\multirow{6}{*}{0.9800 (Acc) \cite{di2020mutual}}&\multirow{6}{*}{\xmark}\\ 
& LGBM&\textbf{1.0000}&\textbf{1.0000}&\textbf{1.0000}&1.2089&\\ 
& XGBoost&0.9722&0.9663&1.0000&1.2133&\\ 
& TabNet&0.8333&0.8497&0.9503&0.7306&\\
& TabPFN&0.9800&0.9785&---&0.9704&\\ 
& TabLM &0.9722&0.9761&1.0000&\textbf{1.2147}&\\ 
\hline
\multirow{6}{*}{Diabetes} & SVM (RBF)&\textbf{0.7662}&0.7411&0.8044&0.4833&\multirow{6}{*}{0.7879 (Acc) \cite{sarkar2022xbnet}}&\multirow{6}{*}{\xmark}\\ 
& LGBM&0.7532&0.7334&0.8129&0.4671&\\ 
& XGBoost&0.7597&0.7301&0.8235&0.4640&\\ 
& TabNet&0.7273&0.6250&\textbf{0.8525}&0.4329&\\
& TabPFN&\textbf{0.7662}&\textbf{0.7433}&0.8211&\textbf{0.4870}&\\ 
& TabLM &0.6423&0.6594&0.6593&0.3962&\\ 
\hline
\multirow{6}{*}{Titanic$\heartsuit$} & SVM (RBF)&0.7765&0.7687&0.8654&0.5376&\multirow{6}{*}{0.7985 (Acc) \cite{sarkar2022xbnet}}&\multirow{6}{*}{\cmark}\\ 
& LGBM&0.7877&0.7747&\textbf{0.8995}&0.5572&\\ 
& XGBoost&0.7989&\textbf{0.7889}&0.8958&0.5812&\\ 
& TabNet&\textbf{0.8212}&0.7612&0.8938&\textbf{0.6192}&\\
& TabPFN&0.8101&0.7344&0.4747&0.5923&\\ 
& TabLM &\textbf{0.8212}& 0.7777&0.8521&0.6001&\\ 
\bottomrule
\end{tabular}
\end{adjustbox}
\end{sc}
\end{small}
\end{center}
\vskip -0.1in
\end{table*}

\paragraph{Handling Missing Data \& Imputation} Lastly in our experiments regarding missing data handling, we evaluated a baseline model that ignores missing values by not serializing any text (equivalent to dropping that cell in tabular data). We then tested two strategies for imputing filler sentences into serialized data. The first strategy (Model: Impute 1) used a sentence that had no relevance to the classification objective, while the second (Model: Impute 2) used a filler sentence related to the classification objectives. We analyzed the differences denoted as $\Delta$ in the logirthm of odds (logits) by subtracting the logits of the two imputed models from the baseline logits to assess how the logits for each class were affected by the imputation. Logits centered at the origin (0,0) indicated that they were typically unaltered, whereas logits that deviated from the origin were heavily altered.

\begin{figure}[h!]
    \centering
    \includegraphics[width=2.5in]{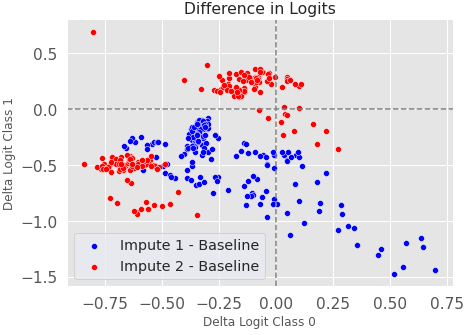}
    \caption{\textbf{Logarithm of Odds Sensitivity Analysis:} We analyzed the logarithm of odds (logits) by computing the differences between the imputed model logits and the baseline logits to determine how much each sample in the test set deviated from the original logit value. We observed that both imputation strategies significantly altered the logit values, demonstrating the sensitivity of these strategies towards the raw probabilites.}
    \label{delta}
    \vspace{-0.4cm}
\end{figure}

Our results, displayed in Figure \ref{delta}, reveal that imputing sentences similar to those seen in \cite{lee2024multimodal} should be used with caution, as they appear to cause significant changes in $\Delta$, potentially altering the final class prediction. This could lead to learning the distribution of the imputed data, which can greatly affect performance, particularly if there is a substantial amount of missing data within a specific feature.

\begin{table*}[h!]
\caption{The benchmarking metrics displayed with Accuracy, F1, Area under the receiver operator curve (AUROC), and Mathews Correlation Coefficent (MCC) scores. * inidcates that the model was not optimized for large tabular tasks. See footnote \ref{fn:sample}. as to why we didn't run the analysis on particular datasets and further details in \cite{hollmann2022tabpfn}. In Heloc dataset we train only on 1000 samples.}
\vskip 0.15in
\begin{center}
\label{evaluation}
\begin{small}
\begin{sc}
\begin{adjustbox}{width=\textwidth}
\begin{tabular}{p{2cm}p{2.8cm}|cccc|c|c}
\toprule
\multicolumn{7}{c}{\textbf{State of the Art Evaluation - Experimental Datasets}} \\ 
Dataset & Method & Accuracy & F1 & AUROC & MCC &  Current State of the Art & TabLm SOTA \\
\hline
\multirow{6}{*}{HELOC $\dagger$} & SVM (RBF)&0.7223&0.7207&0.7903&0.4426&\multirow{6}{*}{N/A} &\multirow{6}{*}{\xmark}\\ 
& LGBM&\textbf{0.7280}&\textbf{0.7267}&0.7958&\textbf{0.4541}&\\ 
& XGBoost&0.7170&0.7157&0.7746&0.4321&\\ 
& TabNet&0.7275&0.7070&\textbf{0.7966}&0.4532&\\
& TabPFN&0.7500*&0.7253*&0.4519*&0.5014*&\\ 
& TabLM &0.7157&0.7025&0.7939&0.4331&\\ 
\hline
\multirow{6}{*}{Fraud $\diamondsuit$} & SVM (RBF)&0.9983&0.4996&0.4790&0.0000&\multirow{6}{*}{0.9530 (AUROC) \cite{xu2023deep}}&\multirow{6}{*}{\xmark}\\ 
& LGBM&0.9994&0.9075&0.9083&0.8167&\\ 
& XGBoost&\textbf{0.9996}&\textbf{0.9293}&\textbf{0.9811}&\textbf{0.8635}&\\ 
& TabNet&0.9994&0.8218&0.9640&0.8215&\\
& TabPFN*&---&---&---&---&\\ 
& TabLM &0.9988&0.9211&0.9155&0.8545&\\ 
\hline
\multirow{6}{*}{Crime$\clubsuit$} & SVM (RBF)&0.2006&0.0088&0.4849&0.2310&\multirow{6}{*}{N/A}&\multirow{6}{*}{\cmark}\\ 
& LGBM&0.2636&\textbf{0.0764}&0.6291&0.2395&\\ 
& XGBoost&0.2606&0.0756&0.6467&0.2389&\\ 
& TabNet&0.3087&0.0502&\textbf{0.7193}&0.2097&\\
& TabPFN*&---&---&---&---&\\ 
& TabLM &\textbf{0.3212}&0.0671&0.6789&\textbf{0.2437}&\\ 
\hline
\multirow{6}{*}{Cancer$\spadesuit$} & SVM (RBF)&1.0000&1.0000&1.0000&1.1428&\multirow{6}{*}{N/A}&\multirow{6}{*}{\xmark}\\ 
& LGBM&1.0000&1.0000&1.0000&1.1428&\\ 
& XGBoost&1.0000&1.0000&1.0000&1.1428&\\ 
& TabNet&0.9814&0.9735&0.9994&0.9749&\\
& TabPFN*&---&---&---&---&\\ 
& TabLM &0.9833&0.9826&0.9864&0.9792&\\ 

\bottomrule
\end{tabular}
\end{adjustbox}
\end{sc}
\end{small}
\end{center}
\vskip -0.1in
\end{table*}

\subsection{SOTA Benchmark}

Having identified the preprocessing steps that are generally beneficial to language models and text serialization, we now proceed with a comprehensive benchmark across all baseline and experimental datasets. This benchmark compares our TabLM against traditional ML algorithms and two specific algorithms from recent literature: Tabnet \cite{arik2021tabnet} and TabPFN \footnote{\label{fn:sample}WARNING: TabPFN is not suitable for datasets with training sizes above 1024 and feature sizes above 10. Predictions become slower and less reliable as dataset size increases. The authors advise against using TabPFN for datasets with over 10k samples due to potential machine crashes from quadratic memory scaling. Consequently, we do not include evaluations as a result on the Crime, Cancer, and Fraud classification datasets.}\cite{hollmann2022tabpfn}. We also include current state-of-the-art models identified by competitions and the open web to showcase what the actual highest metric is. To this end, we introduce a separate column in our benchmarks to highlight these methods as well as their winning performance metric.

\section{Discussion}

\subsection{Language Models benefit from Feature Selection}

From our study on data curation, we identified that among the three techniques, feature selection was the only beneficial data curation strategy. Other strategies, such as feature scaling and handling missing data, showed negative or nuanced results, suggesting that their inclusion could lead to adverse outcomes. Therefore, based on our findings, we advise researchers who use language models on tabular tasks to apply these data curation techniques with caution. We therefore believe more work has to be done in identifying appropriate serialization strategies.

\subsection{Serialization Sensitivity}
\label{sensitivity}

Previous studies \cite{hegselmann2023tabllm} and our experiments with imputation indicate that the logarithm of odds is highly sensitive to minor modifications in the serialized text. Hegselmann et al. found that list readouts and text templates were the most effective serialization strategies. However, our analysis suggests that engineering the input text could significantly enhance or reduce the performance of various language models in classification tasks.

\vspace{-0.05cm}
\subsection{When do I use LM For Tabular tasks?}

\textbf{From this evaluation, it is not conclusively evident that traditional ML techniques or neural network models designed for tabular tasks should be replaced by emerging language model (LM) techniques.} These language models were not optimized for  tabular tasks, and it appears challenging to fine-tune these models without large datasets. This is evident in our baseline experiments where all the datasets had sample sizes less than 1000. This situation is analogous to other deep learning methodologies that require substantial data to tune the large number of parameters and are at risk of overfitting to the training set. Regarding the experimental datasets with larger sample sizes, it also appears that pre-training and transfer learning offer little benefit to these tasks and do not enhance predictive performance.

Therefore, while our TabLM model reached SOTA accuracy levels for specific tasks, other methodologies often yielded more robust results across the board. This finding suggests that these models may not be universally suitable for tabular tasks. However, these models were still competitive, despite not always achieving SOTA performance levels. Extensive research is ongoing to optimize LMs and more recently LLMs for performing tasks on structured data. However, we believe that pre-trained language models should not replace conventional models, and we support the notion that traditional models are still better suited for tabular tasks than deep learning methods \citep{grinsztajn2022tree}.

\section{Conclusion}

In this study, we conducted a series of experiments related to text serialization and compared them to traditional machine learning paradigms. We assessed how various preprocessing steps could enhance or diminish the performance of models. We also performed an benchmarking evaluation against traditional ML models, and two tabular deep learning models and found that pre-trained language models are not better than these exisiting methods. We therefore conclude that pre-trained models are not better than gradient boosted methods.

\paragraph{Code and Data} All code can be found in the \href{https://github.com/Simonlee711/Serialization_SOTA}{Github}. All data is in Appendix Section \ref{dataset}.

\subsection{Impact Statement}

This work aims to advance Tabular Machine Learning by comparing modern NLP language models (LMs) with traditional paradigms. While not covering all aspects of text serialization and tabular characteristics, the study reveals a generally analogous behavior across the evaluated models.

\bibliography{example_paper}
\bibliographystyle{icml2024}

\newpage
\appendix
\onecolumn
\section*{Appendix}
\label{sec:appendix}

In the appendix we cover the following sections:
\begin{itemize}
\vspace{-0.2cm}
    \item Section \ref{supp}: Supplementary Section
    \vspace{-0.2cm}
    \item Section \ref{dataset}: Datasets
    \vspace{-0.2cm}
    \item Section \ref{FMT}: Foundation Models Table
    \vspace{-0.2cm}
    \item Section \ref{hyperparemters}: Hyperparameters of ML Models
    \vspace{-0.2cm}
    \item Section \ref{FSI}: Feature Selection Methods
    \vspace{-0.2cm}
    \item Section \ref{metricss}: Metrics
\end{itemize}

\section{Supplementary Section}
\label{supp}

\begin{figure}[h!]
    \centering
    \includegraphics[width=5in]{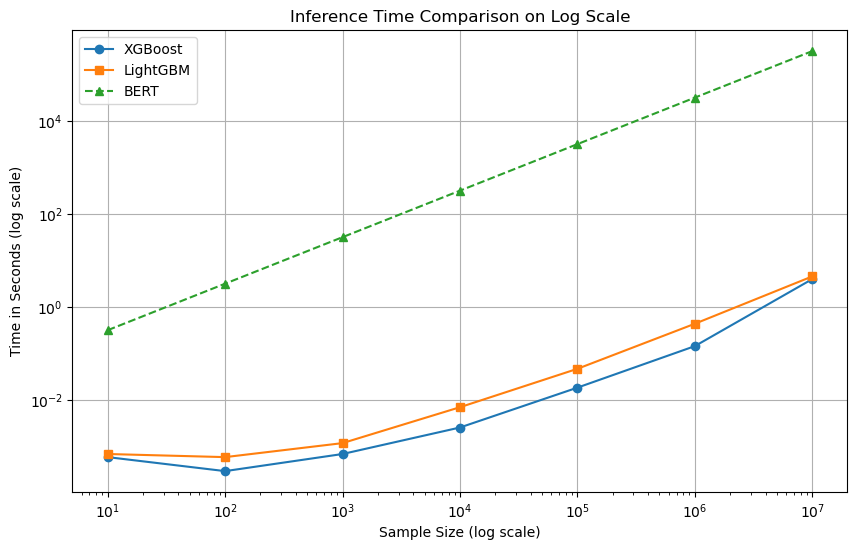}
    \caption{\textbf{Inference Time:} We observe significant differences in training times, where boosting methods prove to be far more efficient than current language model technologies. Language models require processing a substantially larger number of parameters and often need GPU support to achieve competitive runtimes.}
    \label{dd}
\end{figure}

\paragraph{Limitations} One notable limitation of language models in tabular tasks is that they are computationally demanding and costly in terms of runtime compared to methods such as SVMs and gradient boosting. A graphic illustrating runtime at inference is shown in Figure \ref{dd}. Another limitation of language model technologies is the accessibility and lack of inclusivity they create due to their computational demands. We acknowledge that not all groups have access to GPU hardware, which represents a significant barrier in this field of work. \textbf{In this study we therefore elected to use small LM over recent large language models (LLM) due to their the ability to run a local instance without the need for advanced hardware, and the reproduciblility of this work. }

Another notable limitation of this work was highlighted in Section \ref{sensitivity}. Specifically, serialized sentences appear to heavily influence the raw prediction probabilities. While the premise of TabLLM \cite{hegselmann2023tabllm} explored this issue, our research, combined with theirs, still leaves lingering questions about the appropriate strategy for text serialization.

\paragraph{Future Works: } Future work should focus on exploring their scalability and performance as the number of parameters increases \cite{kaplan2020scaling}. As LLMs grow in populatity and size, they demonstrate enhanced capabilities in understanding and producing SOTA performance, but this also introduces challenges related to accessibility to computational resources, and model optimization.

\newpage
\section{Datasets}
\label{dataset}
\subsection{Baseline Datasets}
\paragraph{Iris Dataset} The Iris dataset \cite{misc_iris_53} is a classic dataset in the field of machine learning and statistics, often used for benchmarking classification algorithms. It consists of 150 samples divided equally among three species of Iris flowers: Iris setosa, Iris versicolor, and Iris virginica. Each sample in the dataset is described by four features: sepal length, sepal width, petal length, and petal width, all measured in centimeters. These features are used to predict the species of the Iris flower, making it a multiclass classification problem. The dataset is well-balanced, with 50 samples from each species, providing a clear example for exploring and demonstrating the capabilities of various classification techniques, from simple linear models to more complex, nonlinear classifiers.

\textbf{Link: }\href{https://archive.ics.uci.edu/dataset/53/iris}{Iris Dataset} 
\begin{table}[h!]
    \centering
    \caption{Iris Dataset Features}
\begin{adjustbox}{width=5in}
\begin{tabular}{rrrrr}
\toprule
 sepal length (cm) &  sepal width (cm) &  petal length (cm) &  petal width (cm) &  label \\
\midrule
               5.1 &               3.5 &                1.4 &               0.2 &      0 \\
               4.9 &               3.0 &                1.4 &               0.2 &      0 \\
               4.7 &               3.2 &                1.3 &               0.2 &      0 \\
\bottomrule
\end{tabular}
\end{adjustbox}
\end{table}

\begin{mdframed}
\textbf{Serialized Text:}\\
\texttt{The Iris has sepal Length is \textcolor{red}{5.1} centimeters. Sepal width is \textcolor{red}{3.5} centimeters. Petal length is  \textcolor{red}{1.4}  centimeters. Petal width is  \textcolor{red}{0.2}  centimeters.}
\end{mdframed}

\paragraph{Wine Datset} The Wine \cite{misc_wine_109} dataset is a well-regarded dataset in the machine learning community, commonly used to evaluate multiclass classification algorithms. It comprises 178 instances from three different types of Italian wine: Barolo, Grignolino, and Barbera, derived from the Piedmont region. The dataset is characterized by thirteen attributes, including alcohol, malic acid, ash, alkalinity of ash, magnesium, total phenols, flavanoids, nonflavanoid phenols, proanthocyanins, color intensity, hue, OD280/OD315 of diluted wines, and proline. These attributes are chemically significant and contribute to differentiating one type of wine from another. The objective is to classify each wine into one of the three categories based on its chemical makeup, making it a typical example of a multiclass classification problem. 

\textbf{Link: }\href{https://archive.ics.uci.edu/dataset/109/wine}{Wine Dataset} 

\begin{table}[h!]
    \centering
    \caption{Wine Dataset Features}
\begin{adjustbox}{width=\textwidth}
\begin{tabular}{rrrrrrrrrrrrrr}
\toprule
 alcohol &  malic\_acid &  ash &  alcalinity\_of\_ash &  magnesium &  total\_phenols &  flavanoids &  nonflavanoid\_phenols &  proanthocyanins &  color\_intensity &  hue &  od280/od315\_of\_diluted\_wines &  proline &  label \\
\midrule
    14.2 &         1.7 &  2.4 &               15.6 &      127.0 &            2.8 &         3.1 &                   0.3 &              2.3 &              5.6 &  1.0 &                           3.9 &   1065.0 &      0 \\
    13.2 &         1.8 &  2.1 &               11.2 &      100.0 &            2.6 &         2.8 &                   0.3 &              1.3 &              4.4 &  1.1 &                           3.4 &   1050.0 &      0 \\
    13.2 &         2.4 &  2.7 &               18.6 &      101.0 &            2.8 &         3.2 &                   0.3 &              2.8 &              5.7 &  1.0 &                           3.2 &   1185.0 &      0 \\
\bottomrule
\end{tabular}
\end{adjustbox}
\end{table}

\begin{mdframed}
\textbf{Serialized Text:}\\
\texttt{My wine has an Alcohol percentage of \textcolor{red}{14.2}\%. The Malic Acid is \textcolor{red}{1.7} grams per liter. Ash is \textcolor{red}{2.4}  grams per liter. Alcalinity of ash is \textcolor{red}{15.6} pH. Magnesium is \textcolor{red}{127} milligrams per liter. Total Phenols is \textcolor{red}{2.8}  milligrams per liter. Flavanoids is \textcolor{red}{3.1} milligrams per liter. Nonflavanoid phenols is \textcolor{red}{0.3} milligrams per liter. Proanthocyanins is \textcolor{red}{2.3} milligrams per liter. Color intensity is \textcolor{red}{5.6}. Hue is \textcolor{red}{1.0}. OD280/OD315 of diluted wines is \textcolor{red}{3.9}. Proline is \textcolor{red}{1065}."  
}
\end{mdframed}

\begin{table}[h!]
\centering
\begin{tabular}{lc}
\toprule
\textbf{Feature} & \textbf{Score} \\
\midrule
Alcohol & 99.18 \\
Malic Acid & 33.47 \\
Ash & 11.16 \\
Alkalinity of Ash & 28.68 \\
Magnesium & 5.52 \\
Total Phenols & 78.24 \\
Flavanoids & 272.00 \\
Nonflavanoid Phenols & 26.65 \\
Proanthocyanins & 25.28 \\
Color Intensity & 101.34 \\
Hue & 85.70 \\
OD280/OD315 of Diluted Wines & 175.80 \\
Proline & 151.48 \\
\bottomrule
\end{tabular}
\caption{Importance Scores of Features in the Wine Dataset using ANOVA F-test. We set a threshold of 30 for this problem and removed 4 features in the feature selection.}
\label{tab:feature_scores}
\end{table}

\begin{mdframed}
\newpage
\textbf{Feature Selected Serialized Text:}\\
\texttt{My wine has an Alcohol percentage of \textcolor{red}{14.2}\%. The Malic Acid is \textcolor{red}{1.7} grams per liter. Ash is \textcolor{red}{2.4}  grams per liter. Total Phenols is \textcolor{red}{2.8}  milligrams per liter. Flavanoids is \textcolor{red}{3.1} milligrams per liter. Color intensity is \textcolor{red}{5.6}. Hue is \textcolor{red}{1.0}. OD280/OD315 of diluted wines is \textcolor{red}{3.9}. Proline is \textcolor{red}{1065}."  
}
\end{mdframed}

\paragraph{Diabetes Dataset} The Diabetes dataset \cite{smith1988using}, often referred to as the Pima Indians Diabetes Database, is a frequently used dataset in the domain of medical informatics for predicting the onset of diabetes based on diagnostic measures. This dataset consists of 768 instances, each representing a female at least 21 years old of Pima Indian heritage. The dataset encompasses several medical predictor variables including the number of pregnancies, plasma glucose concentration, diastolic blood pressure, triceps skinfold thickness, 2-hour serum insulin, body mass index, diabetes pedigree function, and age. The target variable indicates whether the individual was diagnosed with diabetes (1) or not (0), making it a binary classification problem. This dataset is pivotal in the development and testing of predictive models aimed at diagnosing diabetes early and has been instrumental in numerous studies related to machine learning in healthcare.

\textbf{Link: }\href{https://www.kaggle.com/datasets/mathchi/diabetes-data-set}{Diabetes Dataset} 

\begin{table}[h!]
    \centering
    \caption{Diabetes Dataset Features}
\begin{adjustbox}{width=\textwidth}
\begin{tabular}{rrrrrrrrr}
\toprule
 Pregnancies &  Glucose &  BloodPressure &  SkinThickness &  Insulin &  BMI &  DiabetesPedigreeFunction &  Age &  Outcome \\
\midrule
           6 &      148 &             72 &             35 &        0 & 33.6 &                       0.6 &   50 &        1 \\
           1 &       85 &             66 &             29 &        0 & 26.6 &                       0.4 &   31 &        0 \\
           8 &      183 &             64 &              0 &        0 & 23.3 &                       0.7 &   32 &        1 \\
\bottomrule
\end{tabular}
\end{adjustbox}
\end{table}

\begin{mdframed}
\textbf{Serialized Text:}\\
\texttt{The Age is \textcolor{red}{50}. The Number of times pregnant is \textcolor{red}{6}. The Diastolic blood pressure is \textcolor{red}{72}. The Triceps skin fold thickness is \textcolor{red}{32}. The Plasma glucose concentration at 2 hours in an oral glucose tolerance test (GTT) is \textcolor{red}{148}. The 2-hour serum insulin is \textcolor{red}{0}. The Body mass index is \textcolor{red}{33.6}. The Diabetes pedigree function is \textcolor{red}{0.6}.}"
\end{mdframed}

\begin{table}[h!]
\centering
\begin{tabular}{lc}
\toprule
\textbf{Feature} & \textbf{Importance Score} \\
\midrule
Pregnancies & 23.93 \\
Glucose & 163.60 \\
Blood Pressure & 2.04 \\
Skin Thickness & 4.80 \\
Insulin & 8.92 \\
BMI & 62.25 \\
Diabetes Pedigree Function & 16.77 \\
Age & 37.07 \\
\bottomrule
\end{tabular}
\caption{Feature Importance Scores in the Diabetes Dataset. We thresholded values that were less than 10 removing 3 features.}
\label{tab:diabetes_feature_importance}
\end{table}
\newpage
\begin{mdframed}
\textbf{Feature Selected Serialized Text:}\\
\texttt{The Age is \textcolor{red}{50}. The Number of times pregnant is \textcolor{red}{6}. The Plasma glucose concentration at 2 hours in an oral glucose tolerance test (GTT) is \textcolor{red}{148}. The Body mass index is \textcolor{red}{33.6}. The Diabetes pedigree function is \textcolor{red}{0.6}.}"
\end{mdframed}

\paragraph{Titanic Dataset} The Titanic dataset \cite{eaton1995titanic} is one of the most iconic datasets used in the realm of data science, especially for beginners practicing classification techniques. It comprises passenger records from the tragic maiden voyage of the RMS Titanic in 1912. This dataset typically includes 891 instances, representing a subset of the total passenger list. Each instance includes various attributes such as passenger class (Pclass), name, sex, age, number of siblings/spouses aboard (SibSp), number of parents/children aboard (Parch), ticket number, fare, cabin number, and port of embarkation. The primary objective with this dataset is to predict a passenger’s survival (1 for survived, 0 for did not survive), making it a binary classification problem. The Titanic dataset not only challenges model builders to predict survival outcomes accurately but also provides an opportunity to explore data preprocessing techniques like handling missing values, feature engineering, and categorical data encoding. It serves as a practical introduction to machine learning tasks and is frequently used in educational settings to demonstrate the steps involved in the data science workflow from preprocessing to model evaluation.

\textbf{Link: }\href{https://www.openml.org/search?type=data&sort=runs&id=40945&status=active}{Titanic Dataset} 

\begin{table}[h!]
    \centering
    \caption{Titanic Dataset Features}
\begin{adjustbox}{width=\textwidth}
\begin{tabular}{rrrllrrrlrll}
\toprule
 PassengerId &  Survived &  Pclass &                                               Name &    Sex &  Age &  SibSp &  Parch &           Ticket &  Fare & Cabin & Embarked \\
\midrule
           1 &         0 &       3 &                            Braund, Mr. Owen Harris &   male & 22.0 &      1 &      0 &        A/5 21171 &   7.2 &   NaN &        S \\
           2 &         1 &       1 & Cumings, Mrs. John Bradley (Florence Briggs Tha... & female & 38.0 &      1 &      0 &         PC 17599 &  71.3 &   C85 &        C \\
           3 &         1 &       3 &                             Heikkinen, Miss. Laina & female & 26.0 &      0 &      0 & STON/O2. 3101282 &   7.9 &   NaN &        S \\
\bottomrule
\end{tabular}
\end{adjustbox}
\end{table}

\begin{mdframed}
\textbf{Serialized Text:}\\
\texttt{Passenger Name is \textcolor{red}{Mr. Own Harris Broaund}. Passenger is \textcolor{red}{22}-years-old. Passenger is \textcolor{red}{male}. They paid \$\textcolor{red}{7.2}. They are in \textcolor{red}{3rd}-class ticket. They embarked from \textcolor{red}{Southhampton}. They are with \textcolor{red}{1} sibling(s)/spouse(s). They are with \textcolor{red}{0} parent(s)/children. They are staying in cabin \textcolor{red}{Unknown}.
}
\end{mdframed}

\begin{mdframed}
\textbf{Modified Serialized Text: \textcolor{red}{(SOTA)}}\\
\texttt{Passenger \textcolor{red}{Mr. Own Harris Broaund}, a \textcolor{red}{22}-year-old \textcolor{red}{male}, paid \$\textcolor{red}{7.2} for a \textcolor{red}{3rd}-class ticket and embarked from \textcolor{red}{Southhampton}. They were accompanied by \textcolor{red}{1} sibling(s)/spouse(s) and \textcolor{red}{0} parent(s)/children, they were aboard in cabin \textcolor{red}{Unknown}.
}
\end{mdframed}

\begin{figure}[h!]
    \centering
    \includegraphics[width=3in]{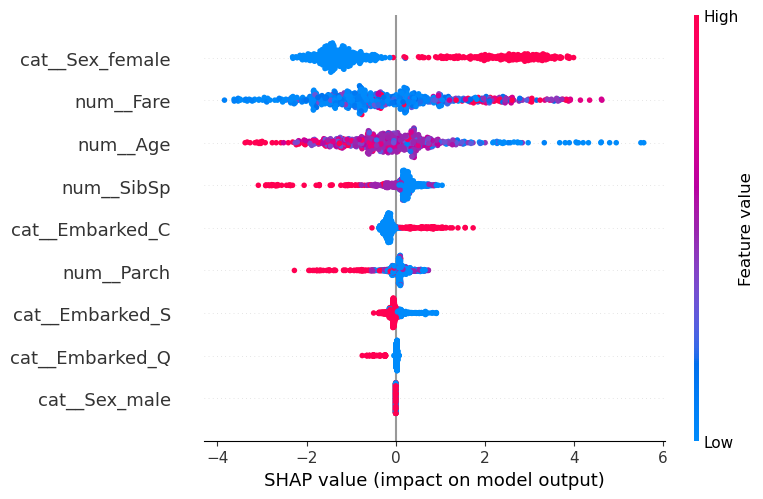}
    \caption{Shap Values from the Titanic Dataset which revealed that Embarked information is minimally contributing to model performance and that women and low aged members (babies), were considered more probable to survive.}
    \label{shap}
\end{figure}

\newpage
\begin{mdframed}
\textbf{Feature Selected Serialized Text:}\\
\texttt{Passenger \textcolor{red}{Mr. Own Harris Broaund}, a \textcolor{red}{22}-year-old \textcolor{red}{male}, paid \$\textcolor{red}{7.2} for a \textcolor{red}{3rd}-class ticket. They were accompanied by \textcolor{red}{1} sibling(s)/spouse(s) and \textcolor{red}{0} parent(s)/children.
}
\end{mdframed}

\subsection{Experimental Datasets}

\paragraph{Home Equity Line of Credit (HELOC) Dataset} The Home Equity Line of Credit (HELOC) dataset is a rich resource for data scientists and machine learning practitioners focusing on financial decision-making processes. This dataset, sourced from real loan applications, includes data from applicants who applied for a home equity line of credit from a lending institution. It features approximately 10,459 instances, each characterized by a series of attributes that are critical in assessing creditworthiness and risk. These attributes include borrower’s credit score, loan to value ratio, number of derogatory remarks, total credit balance, and more, comprising a total of 23 predictive attributes plus a binary target variable. The target variable indicates whether the applicant was approved (1) or rejected (0) for the loan, setting up a binary classification problem. The HELOC dataset not only tests a model's ability to predict loan approval based on complex interactions between various financial indicators but also pushes the boundaries of responsible AI by emphasizing the need for fair and unbiased decision-making systems in finance. It serves as an excellent basis for developing and refining models that deal with imbalanced data, process personal financial information, and require careful feature engineering and selection to predict outcomes accurately.

\textbf{Link: } \href{https://www.kaggle.com/datasets/averkiyoliabev/home-equity-line-of-creditheloc}{HELOC Data}

\begin{table}[h!]
    \centering
    \caption{HELOC Dataset Features Part 1}
    \begin{adjustbox}{width=\textwidth}
    \begin{tabular}{lccccccccc}
    \toprule
    RiskPerformance & ExternalRiskEstimate & MSinceOldestTradeOpen & MSinceMostRecentTradeOpen & AverageMInFile & NumSatisfactoryTrades & NumTrades60Ever2DerogPubRec & NumTrades90Ever2DerogPubRec & PercentTradesNeverDelq & MSinceMostRecentDelq \\
    \midrule
    Bad & 55 & 144 & 4 & 84 & 20 & 3 & 0 & 83 & 2 \\
    Bad & 61 & 58 & 15 & 41 & 2 & 4 & 4 & 100 & -7 \\
    Bad & 67 & 66 & 5 & 24 & 9 & 0 & 0 & 100 & -7 \\
    \bottomrule
    \end{tabular}
    \end{adjustbox}
\end{table}

\begin{table}[h!]
    \centering
    \caption{HELOC Dataset Features Part 2}
    \begin{adjustbox}{width=\textwidth}
    \begin{tabular}{cccccccccc}
    \toprule
    MaxDelq2PublicRecLast12M & MaxDelqEver & NumTotalTrades & NumTradesOpeninLast12M & PercentInstallTrades & MSinceMostRecentInqexcl7days & NumInqLast6M & NumInqLast6Mexcl7days & NetFractionRevolvingBurden & NetFractionInstallBurden \\
    \midrule
    3 & 5 & 23 & 1 & 43 & 0 & 0 & 0 & 33 & -8 \\
    0 & 8 & 7 & 0 & 67 & 0 & 0 & 0 & 0 & -8 \\
    7 & 8 & 9 & 4 & 44 & 0 & 4 & 4 & 53 & 66 \\
    \bottomrule
    \end{tabular}
    \end{adjustbox}
\end{table}
\newpage
\begin{mdframed}
\textbf{Serialized Text:}\\
\texttt{%
External Risk Estimate is \textcolor{red}{55}. 
Months Since Oldest Trade Open is \textcolor{red}{144}. 
Months Since Most Recent Trade Open is \textcolor{red}{4}. 
Average Months In File is \textcolor{red}{84}. 
Number of Satisfactory Trades is \textcolor{red}{20}. 
Number of Trades 60 Ever 2 Derogatory/Public Records is \textcolor{red}{3}. 
Number of Trades 90 Ever 2 Derogatory/Public Records is \textcolor{red}{0}. 
Percent Trades Never Delinquent is \textcolor{red}{83}. 
Months Since Most Recent Delinquency is \textcolor{red}{2}. 
Max Delinquency 2 Public Record Last 12 Months is \textcolor{red}{3}. 
Maximum Delinquency Ever is \textcolor{red}{5}. 
Number of Total Trades is \textcolor{red}{23}. 
Number of Trades Open in Last 12 Months is \textcolor{red}{1}. 
Percent Installment Trades is \textcolor{red}{43}. 
Months Since Most Recent Inquiry Excluding Last 7 Days is \textcolor{red}{0}. 
Number of Inquiries Last 6 Months is \textcolor{red}{0}. 
Number of Inquiries Last 6 Months Excluding Last 7 Days is \textcolor{red}{0}. 
Net Fraction Revolving Burden is \textcolor{red}{33}. 
Net Fraction Installment Burden is \textcolor{red}{-8}. 
Number of Revolving Trades with Balance is \textcolor{red}{8}. 
Number of Installment Trades with Balance is \textcolor{red}{1}. 
Number of Bank/National Trades with High Utilization is \textcolor{red}{1}. 
Percent of Trades with Balance is \textcolor{red}{69}.
}
\end{mdframed}

\begin{table}[h!]
\centering
\caption{Feature Importance Scores for HELOC Dataset. We set the threshold to 50 and remove 8 features.}
\begin{tabular}{lc}
\toprule
\textbf{Feature} & \textbf{Importance Score} \\
\midrule
External Risk Estimate & 390.94 \\
Months Since Oldest Trade Open & 282.23 \\
Months Since Most Recent Trade Open & 14.51 \\
Average Months In File & 371.41 \\
Number of Satisfactory Trades & 113.51 \\
Number of Trades 60 Ever 2 Derog/Public Rec & 45.44 \\
Number of Trades 90 Ever 2 Derog/Public Rec & 20.50 \\
Percent Trades Never Delinquent & 116.84 \\
Months Since Most Recent Delinquency & 33.35 \\
Max Delinquency 2 Public Rec Last 12 Months & 98.07 \\
Max Delinquency Ever & 96.19 \\
Number of Total Trades & 64.18 \\
Number of Trades Open in Last 12 Months & 10.90 \\
Percent Installment Trades & 116.30 \\
Months Since Most Recent Inquiry excl 7 days & 103.23 \\
Number of Inquiries Last 6 Months & 65.35 \\
Number of Inquiries Last 6 Months excl 7 days & 58.71 \\
Net Fraction Revolving Burden & 811.45 \\
Net Fraction Installment Burden & 67.57 \\
Number of Revolving Trades with Balance & 19.75 \\
Number of Installment Trades with Balance & 13.88 \\
Number of Bank/National Trades with High Utilization & 6.33 \\
Percent of Trades with Balance & 337.51 \\
\bottomrule
\end{tabular}
\label{tab:feature_importance}
\end{table}

\begin{mdframed}
\textbf{Feature Selected Serialized Text:}\\
\texttt{%
External Risk Estimate is \textcolor{red}{55}. 
Months Since Oldest Trade Open is \textcolor{red}{144}. 
Average Months In File is \textcolor{red}{84}. 
Number of Satisfactory Trades is \textcolor{red}{20}. 
Percent Trades Never Delinquent is \textcolor{red}{83}. 
Max Delinquency 2 Public Record Last 12 Months is \textcolor{red}{3}. 
Maximum Delinquency Ever is \textcolor{red}{5}. 
Number of Total Trades is \textcolor{red}{23}. 
Percent Installment Trades is \textcolor{red}{43}. 
Months Since Most Recent Inquiry Excluding Last 7 Days is \textcolor{red}{0}. 
Number of Inquiries Last 6 Months is \textcolor{red}{0}. 
Number of Inquiries Last 6 Months Excluding Last 7 Days is \textcolor{red}{0}. 
Net Fraction Revolving Burden is \textcolor{red}{33}. 
Net Fraction Installment Burden is \textcolor{red}{-8}. 
Percent of Trades with Balance is \textcolor{red}{69}.
}
\end{mdframed}

\newpage

\paragraph{Credit Card Fraud Dataset} The Credit Card Fraud dataset \cite{dal2014learned, dal2017credit, dal2015calibrating}, available on Kaggle, is a critical dataset in the financial sector for the development and testing of anomaly detection systems. This dataset contains transactions made by credit cards in September 2013 by European cardholders. It consists of 284,807 transactions, where each transaction is represented by 31 features. These features include 28 numerical input variables (V1 to V28) which are the result of a Principal Component Analysis (PCA) transformation to protect sensitive information, the transaction amount (Amount), and the time since the first transaction in the dataset (Time). The target variable is binary, indicating fraud ('1') or not fraud ('0'), making it a binary classification problem. The dataset is highly imbalanced, with fraud transactions making up only 0.172\% of all transactions. This dataset challenges researchers to effectively detect fraudulent transactions in a highly imbalanced data setting, which is crucial for preventing financial losses due to fraud and is extensively used in machine learning research focused on fraud detection.

\textbf{Link: }\href{https://www.kaggle.com/datasets/mlg-ulb/creditcardfraud/data}{Fraud Dataset}

\begin{table}[h!]
\centering
\caption{Credit Card Transactions Features Part 1}
\begin{tabular}{rrrrrrrrrrrrrrrr}
\toprule
Time & V1 & V2 & V3 & V4 & V5 & V6 & V7 & V8 & V9 & V10 & V11 & V12 & V13 & V14 & V15 \\
\midrule
0.0 & -1.4 & -0.1 & 2.5 & 1.4 & -0.3 & 0.5 & 0.2 & 0.1 & 0.4 & 0.1 & -0.6 & -0.6 & -1.0 & -0.3 & 1.5 \\
0.0 & 1.2 & 0.3 & 0.2 & 0.4 & 0.1 & -0.1 & -0.1 & 0.1 & -0.3 & -0.2 & 1.6 & 1.1 & 0.5 & -0.1 & 0.6 \\
1.0 & -1.4 & -1.3 & 1.8 & 0.4 & -0.5 & 1.8 & 0.8 & 0.2 & -1.5 & 0.2 & 0.6 & 0.1 & 0.7 & -0.2 & 2.3 \\
\bottomrule
\end{tabular}
\end{table}

\begin{table}[h!]
\centering
\caption{Credit Card Transactions Features Part 2}
\begin{tabular}{rrrrrrrrrrrrrrrr}
\toprule
V16 & V17 & V18 & V19 & V20 & V21 & V22 & V23 & V24 & V25 & V26 & V27 & V28 & Amount & Class \\
\midrule
-0.5 & 0.2 & 0.0 & 0.4 & 0.3 & -0.0 & 0.3 & -0.1 & 0.1 & 0.1 & -0.2 & 0.1 & -0.0 & 149.6 & 0 \\
0.5 & -0.1 & -0.2 & -0.1 & -0.1 & -0.2 & -0.6 & 0.1 & -0.3 & 0.2 & 0.1 & -0.0 & 0.0 & 2.7 & 0 \\
-2.9 & 1.1 & -0.1 & -2.3 & 0.5 & 0.2 & 0.8 & 0.9 & -0.7 & -0.3 & -0.1 & -0.1 & -0.1 & 378.7 & 0 \\
\bottomrule
\end{tabular}
\end{table}

\begin{mdframed}
\textbf{Serialized Transaction Data:}\\
\texttt{%
V1 is \textcolor{red}{-1.4}. V2 is \textcolor{red}{-0.1}. V3 is \textcolor{red}{2.5}. V4 is \textcolor{red}{1.4}. V5 is \textcolor{red}{-0.3}. V6 is \textcolor{red}{0.5}. V7 is \textcolor{red}{0.2}. V8 is \textcolor{red}{0.1}. V9 is \textcolor{red}{0.4}. V10 is \textcolor{red}{0.1}. V11 is \textcolor{red}{-0.6}. V12 is \textcolor{red}{-0.6}. V13 is \textcolor{red}{-1.0}. V14 is \textcolor{red}{-0.3}. V15 is \textcolor{red}{1.5}.
V16 is \textcolor{red}{-0.5}. V17 is \textcolor{red}{0.2}. V18 is \textcolor{red}{0.0}. V19 is \textcolor{red}{0.4}. V20 is \textcolor{red}{0.3}, V21 is \textcolor{red}{-0.0}. V22 is \textcolor{red}{0.3}. V23 is \textcolor{red}{-0.1}. V24 is \textcolor{red}{0.1}. V25 is \textcolor{red}{0.1}. V26 is \textcolor{red}{-0.2}. V27 is \textcolor{red}{0.1}. V28 is \textcolor{red}{-0.0}.
}
\end{mdframed}

\begin{table}[h!]
\centering
\caption{Credit Card Transaction Feature Importance Scores. We remove features that are less than 100.}
\begin{tabular}{lc}
\toprule
\textbf{Feature} & \textbf{Importance Score} \\
\midrule
V1 & 2527.72 \\
V2 & 1998.44 \\
V3 & 9026.38 \\
V4 & 4002.88 \\
V5 & 2345.90 \\
V6 & 428.86 \\
V7 & 8861.27 \\
V8 & 87.15 \\
V9 & 2133.98 \\
V10 & 10886.90 \\
V11 & 5309.16 \\
V12 & 15834.84 \\
V13 & 4.13 \\
V14 & 21806.04 \\
V15 & 4.06 \\
V16 & 8917.15 \\
V17 & 27131.19 \\
V18 & 2917.22 \\
V19 & 270.12 \\
V20 & 93.85 \\
V21 & 478.77 \\
V22 & 1.30 \\
V23 & 1.10 \\
V24 & 8.64 \\
V25 & 3.87 \\
V26 & 4.44 \\
V27 & 15.92 \\
V28 & 37.68 \\
Amount & 8.72 \\
\bottomrule
\end{tabular}
\end{table}

\begin{mdframed}
\textbf{Serialized Transaction Data:}\\
\texttt{%
V1 is \textcolor{red}{-1.4}. V2 is \textcolor{red}{-0.1}. V3 is \textcolor{red}{2.5}. V4 is \textcolor{red}{1.4}. V5 is \textcolor{red}{-0.3}. V6 is \textcolor{red}{0.5}. V7 is \textcolor{red}{0.2}. V9 is \textcolor{red}{0.4}. V10 is \textcolor{red}{0.1}. V11 is \textcolor{red}{-0.6}. V12 is \textcolor{red}{-0.6}. V14 is \textcolor{red}{-0.3}.
V16 is \textcolor{red}{-0.5}. V17 is \textcolor{red}{0.2}. V18 is \textcolor{red}{0.0}. V19 is \textcolor{red}{0.4}. V20 is \textcolor{red}{0.3}, V21 is \textcolor{red}{-0.0}.
}
\end{mdframed}

\newpage

\paragraph{San Francisco Crime Dataset} The San Francisco Crime dataset \cite{sf-crime}, available on Kaggle, is an extensive dataset widely used in the domain of predictive modeling and public safety analytics. It includes incidents derived from the San Francisco Police Department's crime incident reporting system, spanning over 12 years from 2003 to 2015. This dataset features over 878,049 instances, each documented with several attributes such as dates, police department district, the category of the crime, the description of the incident, day of the week, and geographical coordinates (latitude and longitude).

The primary objective with this dataset is to predict the category of crime that occurred, making it a multiclass classification problem. Each record is classified into one of 39 distinct crime categories, which include varying offenses from larceny/theft, non-criminal, assault, to drug/narcotic violations. This dataset challenges data scientists to analyze and predict crime patterns based on temporal and spatial features, which is crucial for law enforcement agencies to allocate resources effectively and improve public safety. The San Francisco Crime dataset not only serves as a critical resource for training machine learning models to understand urban crime dynamics but also provides insights into the effectiveness of different policing strategies over time.

\textbf{Link: }\href{https://www.kaggle.com/c/sf-crime}{Crime Dataset}

\begin{table}[h!]
\centering
\caption{Crime Dataset Features}
\begin{adjustbox}{width=1\textwidth} 
\begin{tabular}{lllllllrr}
\toprule
              Dates &       Category &                 Descript & DayOfWeek & PdDistrict &     Resolution &                   Address &      X &    Y \\
\midrule
2015-05-13 23:53:00 &       WARRANTS &           WARRANT ARREST & Wednesday &   NORTHERN & ARREST, BOOKED &        OAK ST / LAGUNA ST & -122.425 & 37.774 \\
2015-05-13 23:53:00 & OTHER OFFENSES & TRAFFIC VIOLATION ARREST & Wednesday &   NORTHERN & ARREST, BOOKED &        OAK ST / LAGUNA ST & -122.425 & 37.774 \\
2015-05-13 23:33:00 & OTHER OFFENSES & TRAFFIC VIOLATION ARREST & Wednesday &   NORTHERN & ARREST, BOOKED & VANNESS AV / GREENWICH ST & -122.424 & 37.800 \\
\bottomrule
\end{tabular}
\end{adjustbox}
\end{table}

\begin{mdframed}
\textbf{Serialized Sentence:}\\
\texttt{The description of the incident was \textcolor{red}{WARRANT ARREST}. The crime happened on \textcolor{red}{Wednesday} in the \textcolor{red}{NORTHERN} police district. The incident happened at \textcolor{red}{OAK ST / LAGUNA ST}, with coordinates (\textcolor{red}{-122.4}, \textcolor{red}{37.8}).\\}
\end{mdframed}

\paragraph{Gene Expression Profiles for Cancer Target Identification Dataset} The Gene Expression Profiles dataset \cite{misc_gene_expression_cancer_rna-seq_401} is a vital resource in the burgeoning field of machine learning for drug discovery, specifically in identifying targets for cancer therapies. This dataset consists of gene expression profiles derived from various cancer patients. It includes data from multiple studies focused on different types of cancer, where each sample is described by potentially thousands of gene expression features, reflecting the activity levels of various genes in the tissues sampled from cancer patients.

The primary objective with this dataset is to distinguish between different cancer types or to predict the response of various cancers to treatments, making it an essential tool for multiclass classification or regression problems in biomedical research. The complexity of the dataset, due to the high dimensionality of the feature space and the biological variability among samples, poses significant challenges in model building, feature selection, and interpretation of results.

\textbf{Link: }\href{https://www.ebi.ac.uk/training/online/courses/machine-learning-drug-discovery/identifying-targets-for-cancer-using-gene-expression-profiles/}{Cancer Dataset}

\begin{table}[h!]
\centering
\caption{Cancer Dataset Features}
\begin{adjustbox}{width=1\textwidth} 
\begin{tabular}{rrrrrrrrrrrrrrrrrrrr}
\toprule
 gene\_0 &  gene\_1 &  gene\_2 &  gene\_3 &  gene\_4 &  gene\_5 &  gene\_6 &  gene\_7 &  gene\_8 &  gene\_9 &  gene\_10 &  gene\_11 &  gene\_12 &  gene\_13 &  gene\_14 &  gene\_15 &  gene\_16 &  gene\_17 &  ... & gene\_20000\\
 \midrule
    0.0 &     2.0 &     3.3 &     5.5 &    10.4 &     0.0 &     7.2 &     0.6 &     0.0 &     0.0 &      0.6 &      1.3 &      2.0 &      0.6 &      0.0 &      0.0 &      0.0 &      0.0 &  ... & 0.4 \\
    0.0 &     0.6 &     1.6 &     7.6 &     9.6 &     0.0 &     6.8 &     0.0 &     0.0 &     0.0 &      0.0 &      0.6 &      2.5 &      1.0 &      0.0 &      0.0 &      0.0 &      0.0 &   ... & 0.0 \\
    0.0 &     3.5 &     4.3 &     6.9 &     9.9 &     0.0 &     7.0 &     0.5 &     0.0 &     0.0 &      0.0 &      0.5 &      2.0 &      1.1 &      0.0 &      0.0 &      0.0 &      0.0 &    ... &1.3 \\
\bottomrule
\end{tabular}
\end{adjustbox}
\end{table}

\begin{mdframed}
\textbf{Serialized Text:}\\
\texttt{%
Gene 0 is \textcolor{red}{0.0}. Gene 1 is \textcolor{red}{0.6}. Gene 2 is \textcolor{red}{1.6}. Gene 3 is \textcolor{red}{7.6}. Gene 4 is \textcolor{red}{9.6}. Gene 5 is \textcolor{red}{0.0}. Gene 6 is \textcolor{red}{6.8}. Gene 7 is \textcolor{red}{0.0}. Gene 8 is \textcolor{red}{0.0}. Gene 9 is \textcolor{red}{0.0}. \\
Gene 10 is \textcolor{red}{0.0}. Gene 11 is \textcolor{red}{0.6}. Gene 12 is \textcolor{red}{2.5}. Gene 13 is \textcolor{red}{1.0}. Gene 14 is \textcolor{red}{0.0}. Gene 15 is \textcolor{red}{0.0}. Gene 16 is \textcolor{red}{0.0}. Gene 17 is \textcolor{red}{0.0}. Gene 18 is \textcolor{red}{0.0}. Gene 19 is \textcolor{red}{11.1}. \\
Gene 20 is \textcolor{red}{3.6}. Gene 21 is \textcolor{red}{0.0}. Gene 22 is \textcolor{red}{10.1}. Gene 23 is \textcolor{red}{0.0}. Gene 24 is \textcolor{red}{0.0}. Gene 25 is \textcolor{red}{0.0}. Gene 26 is \textcolor{red}{9.9}. Gene 27 is \textcolor{red}{8.5}. Gene 28 is \textcolor{red}{1.2}. Gene 29 is \textcolor{red}{4.9}. \\
Gene 30 is \textcolor{red}{0.0}. Gene 31 is \textcolor{red}{0.0}. Gene 32 is \textcolor{red}{5.8}. Gene 33 is \textcolor{red}{1.3}. Gene 34 is \textcolor{red}{13.3}. Gene 35 is \textcolor{red}{6.7}. Gene 36 is \textcolor{red}{0.6}. Gene 37 is \textcolor{red}{0.0}. Gene 38 is \textcolor{red}{9.5}. Gene 39 is \textcolor{red}{0.8}. \\
Gene 40 is \textcolor{red}{9.7}. Gene 41 is \textcolor{red}{0.0}. Gene 42 is \textcolor{red}{0.3}. Gene 43 is \textcolor{red}{0.0}. Gene 44 is \textcolor{red}{2.7}. Gene 45 is \textcolor{red}{6.7}. Gene 46 is \textcolor{red}{9.8}. Gene 47 is \textcolor{red}{8.8}. Gene 48 is \textcolor{red}{11.5}...
}
\end{mdframed}

\begin{figure}[H]
    \centering
    \includegraphics[width=3in]{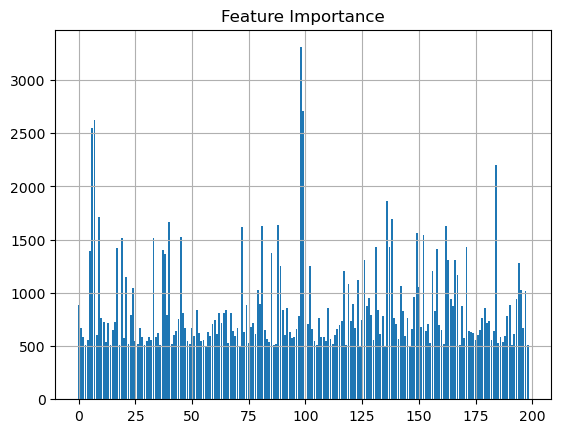}
    \caption{F\textbf{eature Importance:} Assessing the Importance of Genes. Displayed are the ~200 genes with a feature importance of 500 or above.}
    \label{fig:enter-label}
\end{figure}

\subsection{Feature Scaling Experiments}

We performed various feature scaling techniques to correct the skewness of the Titanic data set. Below we display examples of serialized sentences with the applied transforms.

\begin{figure}[H]
    \centering
    \includegraphics[width=2.5in]{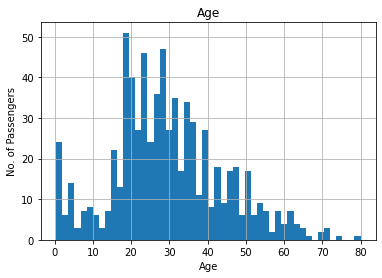}
    \includegraphics[width=2.5in]{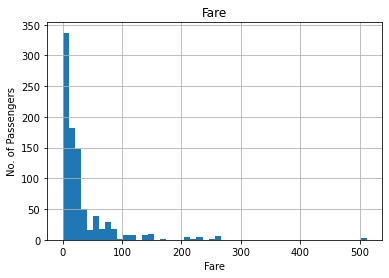}
    \caption{\textbf{Feature Scaling:} Enhancing the Titanic Dataset by Correcting Skewness and Mitigating Outlier Effects}
    \label{fig:enter-label}
\end{figure}

\subsection*{Standardization}

\begin{equation}
z = \frac{x - \mu}{\sigma}
\end{equation}

These symbols denote the following: \( z \) represents the standardized value, \( x \) stands for the original value, \( \mu \) denotes the mean of the data, and \( \sigma \) signifies the standard deviation of the data.

\begin{mdframed}
\textbf{Standardized Selected Serialized Text:}\\
\texttt{Passenger \textcolor{red}{Mr. Own Harris Broaund}, a \textcolor{red}{-0.565}-year-old \textcolor{red}{male}, paid \$\textcolor{red}{-0.502} for a \textcolor{red}{3rd}-class ticket. They were accompanied by \textcolor{red}{1} sibling(s)/spouse(s) and \textcolor{red}{0} parent(s)/children.
}
\end{mdframed}

\subsection*{Normalization}

\begin{equation}
x_{\text{norm}} = \frac{x}{\text{max}(x)}
\end{equation}

In this context, \( x_{\text{norm}} \) denotes the normalized value of \( x \), where \( x \) stands for the original value, and \( \text{max}(x) \) represents the maximum value in the dataset.

\begin{mdframed}
\textbf{Normalized Serialized Text:}\\
\texttt{Passenger \textcolor{red}{Mr. Own Harris Broaund}, a \textcolor{red}{0.271}-year-old \textcolor{red}{male}, paid \$\textcolor{red}{0.014} for a \textcolor{red}{3rd}-class ticket. They were accompanied by \textcolor{red}{1} sibling(s)/spouse(s) and \textcolor{red}{0} parent(s)/children.
}
\end{mdframed}

\subsection*{Log Transformation}
\begin{equation}
y = \log(x)
\end{equation}

In this context, \( y \) represents the logarithmically transformed value of \( x \), where \( x \) stands for the original value.

\begin{mdframed}
\textbf{Feature Selected Serialized Text:}\\
\texttt{Passenger \textcolor{red}{Mr. Own Harris Broaund}, a \textcolor{red}{3.135}-year-old \textcolor{red}{male}, paid \$\textcolor{red}{2.110} for a \textcolor{red}{3rd}-class ticket. They were accompanied by \textcolor{red}{1} sibling(s)/spouse(s) and \textcolor{red}{0} parent(s)/children.
}
\end{mdframed}

\newpage
\section{Foundation Models Table}
\label{FMT}
\begin{table}[H]
\centering
\caption{The foundation models evaluated for becoming the backbone for TabLM.}
\begin{tabular}{p{6cm}p{8cm}}
\toprule
\textbf{Model} & \textbf{Description} \\
\midrule
\textbf{BERT} \cite{devlin2018bert} & Originally pretrained on a corpus consisting of Wikipedia and BookCorpus using masked language modeling (MLM) and next sentence prediction (NSP) tasks to generate bidirectional context representations.\\
\midrule
\textbf{DistilBERT} \cite{sanh2019distilbert} & A lighter version of BERT, retaining most of its predecessor's capabilities but with fewer parameters, pretrained using a knowledge distillation process during the MLM task.\\
\midrule
\textbf{RoBERTa} \cite{liu2019roberta} & A variant of BERT optimized through more extensive training on larger data and removing the NSP task, focusing solely on the MLM for better performance.\\
\midrule
\textbf{Electra} \cite{clark2020electra} & Trained using a replaced token detection rather than masked language modeling, Electra discriminates between "real" and "fake" tokens across a corpus, allowing for more efficient learning.\\
\midrule
\textbf{XLNet} \cite{yang2019xlnet}& Combines the best of autoregressive and autoencoding techniques, pretrained on a permutation-based language modeling task, which captures bidirectional contexts dynamically.\\
\midrule
\textbf{Albert} \cite{lan2019albert} & A lite BERT that introduces parameter-reduction techniques to increase training speed and lower memory consumption, focusing on MLM and sentence-order prediction.\\
\midrule
\textbf{Deberta} \cite{he2020deberta} & Enhances BERT and RoBERTa models by incorporating disentangled attention and a new way of encoding positional information, improving on MLM and NSP tasks.\\
\midrule
\textbf{GPT-2} \cite{radford2019language} & Utilizes a left-to-right autoregressive approach in its pretraining, allowing each token to condition on the previous tokens in a sequence, optimized for a variety of natural language understanding tasks.\\
\midrule
\textbf{Longformer} \cite{beltagy2020longformer} & Designed for longer texts, this model extends the BERT architecture by employing a combination of sliding window and global attention mechanisms, focusing on efficiency and scalability.\\
\midrule
\textbf{GTE Large} \cite{li2023towards} & The general text embedding model (GTE) using a multi-contrastive learning pre-training objective. Scored very high in the MTEB benchmark in Text Classification.\\
\midrule
\textbf{GTE Base} & Similar to GTE Large but with fewer parameters, focused on achieving comparable performance to larger models while being more computationally efficient.\\
\bottomrule
\label{TableModel}
\end{tabular}
\end{table}

\subsection{Results of Language Model Evaluation}
\label{resultsss}
\begin{table}[H]
\centering
\caption{Performance Metrics for Language Models}
\label{tab:metrics}
\begin{tabular}{l|ccccccccc}
\toprule
Model       & Loss    & Accuracy & Precision & Recall & F1 Score & AUROC  & AUPRC  & Runtime (s) & Samples/s \\ \midrule
Bert        & 0.4903  & 0.7821   & 0.7536    & 0.7027 & 0.7273   & 0.8483 & 0.8262 & 5.0933      & 35.144    \\
DistilBert  & \textbf{0.4535}  & 0.8045   & 0.7097    & \textbf{0.8919 }& 0.7904   & \textbf{0.8743} & \textbf{0.8426} & 2.6072      & 68.656    \\
RoBERTa     & 0.5547  & 0.7989   & 0.7317    & 0.8108 & 0.7692   & 0.8206 & 0.7448 & 4.7434      & 37.737    \\
Electra     & 0.4583  & \textbf{0.8268}   & 0.7529    & 0.8649 & \textbf{0.8050}   & 0.8515 & 0.7665 & 5.1101      & 35.029    \\
XLNet       & 0.5574  & 0.7821   & 0.7536    & 0.7027 & 0.7273   & 0.8529 & 0.8222 & 17.336      & 10.325    \\
Albert      & 0.4802  & 0.7989   & 0.7262    & 0.8243 & 0.7722   & 0.8387 & 0.7637 & 5.8252      & 30.729    \\
DeBERTa     & 0.5057  & 0.7933   & 0.7342    & 0.7838 & 0.7582   & 0.8059 & 0.7006 & 3.2567      & 54.964    \\
GPT2        & 0.6947  & 0.6592   & 0.8824    & 0.2027 & 0.3297   & 0.8408 & 0.7877 & \textbf{2.0704}      & \textbf{86.456}    \\
Longformer  & 0.5092  & 0.7989   & 0.7436    & 0.7838 & 0.7632   & 0.8138 & 0.6742 & 3.7726      & 47.447    \\
GTE-large   & 0.5226  & 0.7933   & 0.7761    & 0.7027 & 0.7376   & 0.8704 & 0.7947 & 6.4885      & 27.587    \\
GTE Base    & 0.5336  & 0.7821   & \textbf{0.9070}    & 0.5270 & 0.6667   & 0.8725 & 0.8139 & 2.1677      & 82.575    \\
\bottomrule
\end{tabular}
\end{table}

\vspace{1cm}

\begin{figure}[H]
    \centering
    \includegraphics[width=5in]{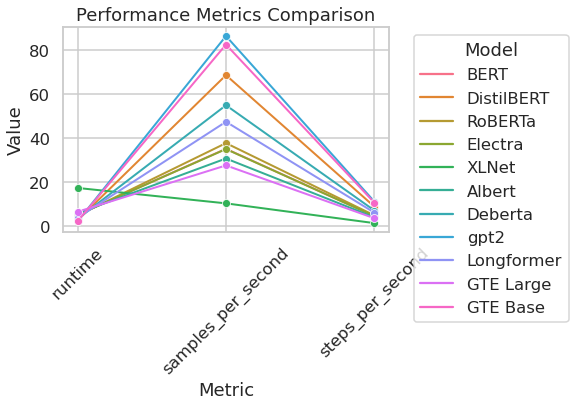}
    \caption{\textbf{Efficiency Evalutation:} Comprehensive Performance Evaluation of Language Models: Analyzing Sample Efficiency, and Runtime Metrics}
    \label{fig:enter-label}
\end{figure}

\newpage
\section{Hyperparameters of Baseline Models}
\label{hyperparemters}

Hyperparameters play a pivotal role in machine learning, governing various aspects of the model training process. In our work, we utilized grid search to systematically explore the optimal settings for different models. We extracted hyperparameters from the \cite{hegselmann2023tabllm} paper for the LGBM and XGBoost mdoels. For XGBoost, we configured parameters such as \texttt{max\_depth} ranging from 2 to 12, \texttt{lambda\_l1} and \texttt{lambda\_l2} from \(1e-8\) to $1.0$, and \texttt{eta} from 0.01 to 0.3. For LightGBM, we examined \texttt{num\_leaves} from 2 to 4096, \texttt{lambda\_l1} and \texttt{lambda\_l2} extending up to 10.0, and \texttt{learning\_rate} from 0.01 to 0.3. The SVM model with an RBF kernel was tested with \texttt{C} values between 0.1 and 100, and \texttt{gamma} values including 0.001 to 1, as well as \texttt{auto} and \texttt{scale}. This comprehensive hyperparameter tuning enhances the model's performance by ensuring the most effective parameter combinations are identified, leading to improved accuracy and robustness.

\renewcommand{\arraystretch}{1.5} 

\begin{tabular}{|>{\centering\arraybackslash}m{6cm}|>{\centering\arraybackslash}m{10cm}|}
\hline
\multicolumn{2}{|c|}{\textbf{XGBoost}} \\
\hline
\textbf{Model} & \texttt{xgb.XGBClassifier(random\_state=42)} \\
\hline
\textbf{Parameters} & \begin{tabular}{@{}c@{}}
                       max\_depth: [2, 4, 6, 8, 10, 12] \\
                       lambda\_l1: [1e-8, 1e-7, 1e-6, 1e-5, 1e-4, 1e-3, 1e-2, 1e-1, 1.0] \\
                       lambda\_l2: [1e-8, 1e-7, 1e-6, 1e-5, 1e-4, 1e-3, 1e-2, 1e-1, 1.0] \\
                       eta: [0.01, 0.03, 0.1, 0.3]
                     \end{tabular} \\
\hline
\end{tabular}

\vspace{1em} 

\begin{tabular}{|>{\centering\arraybackslash}m{6cm}|>{\centering\arraybackslash}m{10cm}|}
\hline
\multicolumn{2}{|c|}{\textbf{LightGBM}} \\
\hline
\textbf{Model} & \texttt{lgb.LGBMClassifier(random\_state=42)} \\
\hline
\textbf{Parameters} & \begin{tabular}{@{}c@{}}
                       num\_leaves: [2, 4, 8, 16, 32, 64, 128, 256, 512, 1024, 2048, 4096] \\
                       lambda\_l1: [1e-8, 1e-7, 1e-6, 1e-5, 1e-4, 1e-3, 1e-2, 1e-1, 1.0, 10.0] \\
                       lambda\_l2: [1e-8, 1e-7, 1e-6, 1e-5, 1e-4, 1e-3, 1e-2, 1e-1, 1.0, 10.0] \\
                       learning\_rate: [0.01, 0.03, 0.1, 0.3]
                     \end{tabular} \\
\hline
\end{tabular}

\vspace{1em} 

\begin{tabular}{|>{\centering\arraybackslash}m{6cm}|>{\centering\arraybackslash}m{10cm}|}
\hline
\multicolumn{2}{|c|}{\textbf{SVM (RBF)}} \\
\hline
\textbf{Model} & \texttt{SVC(probability=True, random\_state=42)} \\
\hline
\textbf{Parameters} & \begin{tabular}{@{}c@{}}
                       C: [0.1, 1, 10, 100] \\
                       gamma: [0.001, 0.01, 0.1, 1, 'auto', 'scale'] \\
                       kernel: ['rbf']
                     \end{tabular} \\
\hline
\end{tabular}

\newpage

\section{Feature Selection Methods}
\label{FSI}
\label{FS}

\paragraph{ANOVA F-test} The ANOVA F-test feature selection method works by computing the ANOVA F-value between each feature and the target variable for classification tasks. The ANOVA F-value is a ratio of the between-group variability to the within-group variability, and it measures how well a feature can separate the samples into different classes.

Mathematically, the ANOVA F-value for a feature $X$ and a target variable $Y$ with $k$ classes is calculated as follows:

\begin{enumerate}
    \item Calculate the mean of $X$ within each class: $\mu_j = \sum_{Y_i = j} X_i / n_j$, where $n_j$ is the number of samples in class $j$
    \item Calculate the overall mean of $X$: $\mu = \sum_i X_i / n$, where $n$ is the total number of samples

    \item Calculate the between-group sum of squares (SSB): $$\text{SSB} = \sum_j n_j (\mu_j - \mu)^2$$

    \item Calculate the within-group sum of squares (SSW):$$\text{SSW} = \sum_{Y_i = j} (X_i - \mu_j)^2$$
    \item Calculate the ANOVA F-value: $$F = \frac{\text{SSB} / (k - 1)}{\text{SSW} / (n - k)}$$
\end{enumerate}

The higher the F-value, the more discriminative the feature is for separating the classes. 

\paragraph{SHAP Values:} The SHAP (SHapley Additive exPlanations) value is a method to explain the output of an XGBoost model \( f \) for a given input vector \( x = (x_1, x_2, \ldots, x_p) \). The SHAP value \( \phi_j(x) \) for feature \( j \) and instance \( x \) is calculated as:

$$\phi_j(x) = \sum_{S \subseteq N \setminus \{j\}} \frac{|S|!(|N|-|S|-1)!}{|N|!}[f_{x}(S \cup \{j\}) - f_{x}(S)]$$

where:

\begin{itemize}
    \item \( N = \{1, 2, \ldots, p\} \) is the set of all feature indices.
    \item \( S \) is a subset of feature indices from \( N \), representing a coalition of features.
    \item \( f_{x}(S) \) is the prediction of the model \( f \) for instance \( x \) using only the features indexed by \( S \).
    \item \( |S| \) is the cardinality (number of elements) of the set \( S \).
\end{itemize}

The SHAP value \( \phi_j(x) \) represents the weighted average of the marginal contributions of feature \( j \) to the model's prediction, with the weights derived from the Shapley value formulation in cooperative game theory.

Specifically, the term \( [f_{x}(S \cup \{j\}) - f_{x}(S)] \) denotes the marginal contribution of feature \( j \) to the prediction when it is added to the coalition of features \( S \). The weight \( \frac{|S|!(|N|-|S|-1)!}{|N|!} \) is the Shapley value weight, ensuring a fair distribution of the total prediction among the features.

To compute the SHAP values, the XGBoost model needs to be evaluated on all possible subsets of features, which can be computationally intensive for high-dimensional datasets. However, efficient approximation algorithms are available in the SHAP library that estimate the SHAP values with reasonable accuracy.

After computing the SHAP values, they can be used for feature selection by ranking the features based on their average absolute SHAP values or by applying a threshold to identify the most important features.

\newpage
\section{Metrics}

For pedgogical purposes we define the metrics used in the study.
\label{metricss}

\subsection{Binary Classification Metrics}

\paragraph{Accuracy}
   $$ \text{Accuracy} = \frac{TP + TN}{TP + TN + FP + FN} $$
   
Where \( TP \) represents True Positives, \( TN \) represents True Negatives, \( FP \) represents False Positives, and \( FN \) represents False Negatives.

\paragraph{F1 Score}
   $$ \text{F1} = 2 \cdot \frac{\text{Precision} \cdot \text{Recall}}{\text{Precision} + \text{Recall}} $$
   
Where Precision is calculated as \( \frac{TP}{TP + FP} \) and Recall is calculated as \( \frac{TP}{TP + FN} \).

\paragraph{Area Under the Receiver Operating Characteristic Curve (AUROC)}
   $$ \text{AUROC} = \int_{0}^{1} TPR(FPR) \, d(\text{FPR}) $$
   
Where \( TPR \) is the True Positive Rate, calculated as \( \frac{TP}{TP + FN} \), and \( FPR \) is the False Positive Rate, calculated as \( \frac{FP}{FP + TN} \).

\paragraph{Matthews Correlation Coefficient (MCC)}
   $$ \text{MCC} = \frac{TP \cdot TN - FP \cdot FN}{\sqrt{(TP + FP)(TP + FN)(TN + FP)(TN + FN)}} $$

\subsection{Multiclass Classification Metrics}

\paragraph{Accuracy}
   $$ \text{Accuracy} = \frac{\sum_{i} TP_i}{\sum_{i} (TP_i + FP_i + FN_i)} $$
   
Where \( TP_i \) represents True Positives for class \( i \), \( FP_i \) represents False Positives for class \( i \), and \( FN_i \) represents False Negatives for class \( i \).

\paragraph{F1 Score (Macro-Averaged)}
   $$ \text{F1}_{\text{macro}} = \frac{1}{C} \sum_{i=1}^{C} \text{F1}_i $$
 Where \( C \) is the number of classes, \( \text{F1}_i = 2 \cdot \frac{\text{Precision}_i \cdot \text{Recall}_i}{\text{Precision}_i + \text{Recall}_i} \), \( \text{Precision}_i = \frac{TP_i}{TP_i + FP_i} \), and \( \text{Recall}_i = \frac{TP_i}{TP_i + FN_i} \).

\paragraph{Matthews Correlation Coefficient (MCC)}
   $$ \text{MCC} = \frac{\sum_{i} \sum_{j} (TP_{i,j} \cdot TN_{i,j} - FP_{i,j} \cdot FN_{i,j})}{\sqrt{\prod_{i} (TP_{i} + FP_{i})(TP_{i} + FN_{i})(TN_{i} + FP_{i})(TN_{i} + FN_{i})}} $$

Where \( TP_{i,j} \) represents True Positives for classes \( i \) and \( j \), \( TN_{i,j} \) represents True Negatives for classes \( i \) and \( j \), \( FP_{i,j} \) represents False Positives for classes \( i \) and \( j \), and \( FN_{i,j} \) represents False Negatives for classes \( i \) and \( j \).

\end{document}